\documentclass[sn-nature]{sn-jnl}

\usepackage{graphicx}%
\usepackage{multirow}%
\usepackage{amsmath,amssymb,amsfonts}%
\usepackage{amsthm}%
\usepackage{mathrsfs}%
\usepackage[title]{appendix}%
\usepackage{xcolor}%
\usepackage{textcomp}%
\usepackage{manyfoot}%
\usepackage{booktabs}%
\usepackage[ruled,lined,linesnumbered]{algorithm2e}
\usepackage{algpseudocode}
\usepackage{listings}%
\usepackage{subcaption}
\usepackage{longtable}
\usepackage{lipsum}
\usepackage{upgreek}
\usepackage{lineno}

\let\oldnl\nl
\newcommand{\nonl}{\renewcommand{\nl}{\let\nl\oldnl}}

\newcommand{\beginsupplement}{%
        \setcounter{table}{0}
        \renewcommand{\thetable}{S-\arabic{table}}%
        \setcounter{figure}{0}
        \renewcommand{\thefigure}{S-\arabic{figure}}
        \setcounter{section}{0}
        \renewcommand{\thesection}{S-\arabic{section}}%
        \setcounter{page}{1}
        \renewcommand{\thepage}{\arabic{page}}
        \setcounter{equation}{1}
        \renewcommand{\theequation}{S-\arabic{equation}}
     }

\usepackage[resetlabels]{multibib}
\newcites{SM}{References}

\begin{document}

\title[Article Title]{A Self-Supervised Robotic System for Autonomous Contact-Based Spatial Mapping of Semiconductor Properties}






\author*[1]{\fnm{Alexander E.} \sur{Siemenn}}\email{asiemenn@mit.edu}

\author[1]{\fnm{Basita} \sur{Das}}\email{dasb@mit.edu}

\author[1,2]{\fnm{Kangyu} \sur{Ji}}\email{axvcb1597382@gmail.com}

\author[1]{\fnm{Fang} \sur{Sheng}}\email{shengf22@mit.edu}

\author[1]{\fnm{Tonio} \sur{Buonassisi}}\email{buonassisi@mit.edu}

\affil[1]{\orgdiv{Department of Mechanical Engineering}, \orgname{Massachusetts Institute of Technology}, \orgaddress{\street{77 Massachusetts Avenue}, \city{Cambridge}, \postcode{02139}, \state{Massachusetts}, \country{USA}}}
\affil[2]{\orgdiv{Research Laboratory of Electronics}, \orgname{Massachusetts Institute of Technology}, \orgaddress{\street{77 Massachusetts Avenue}, \city{Cambridge}, \postcode{02139}, \state{Massachusetts}, \country{USA}}}

\abstract{

Integrating robotically driven contact-based material characterization techniques into self-driving laboratories can enhance measurement quality, reliability, and throughput. While deep learning models support robust autonomy, current methods lack reliable pixel-precision positioning and require extensive labeled data. To overcome these challenges, we propose an approach for building self-supervised autonomy into contact-based robotic systems that teach the robot to follow domain expert measurement principles at high-throughputs. Firstly, we design a vision-based, self-supervised convolutional neural network (CNN) architecture that uses differentiable image priors to optimize domain-specific objectives, refining the pixel precision of predicted robot contact poses by 20.0\% relative to existing approaches. Secondly, we design a reliable graph-based planner for generating distance-minimizing paths to accelerate the robot measurement throughput and decrease planning variance by 6x. We demonstrate the performance of this approach by autonomously driving a 4-degree-of-freedom robotic probe for 24 hours to characterize semiconductor photoconductivity at 3,025 uniquely predicted poses across a gradient of drop-casted perovskite film compositions, achieving throughputs over 125 measurements per hour. Spatially mapping photoconductivity onto each drop-casted film reveals compositional trends and regions of inhomogeneity, valuable for identifying manufacturing process defects. With this self-supervised CNN-driven robotic system, we enable high-precision and reliable automation of contact-based characterization techniques at high throughputs, thereby allowing the measurement of previously inaccessible yet important semiconductor properties for self-driving laboratories.

}

\keywords{autonomous robotics, pose prediction, self-supervised, spatial differentiability, reliable path planning, high-throughput, contact-based characterization}

\maketitle

\begin{figure}[h!p]
    \begin{center}
    \includegraphics[width=1\columnwidth]{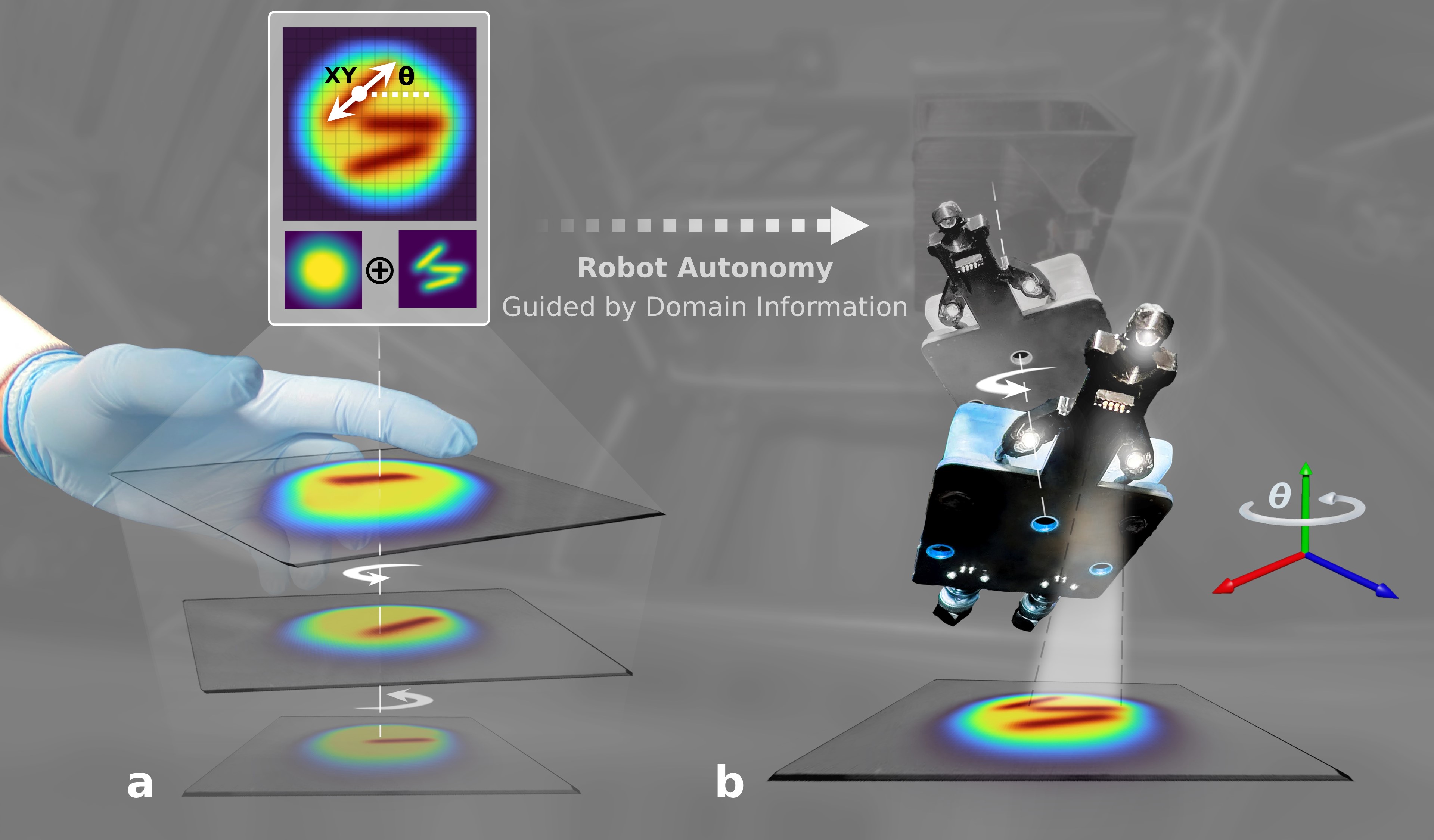}
    \end{center}
       \caption{Using domain information to guide robot autonomy. \textbf{a}, A materials science domain expert chooses the positions and angles of three distinct contact points (red) of a probe to measure a semiconductor film, ensuring maximum coverage while avoiding overlap. Using vision-guided deep learning, this domain information is embedded into the model's objective-based loss functions. \textbf{b}, A 4-degree-of-freedom robotic probe trained using this domain information-embedded deep learning model emulates the measurement procedure of a domain expert autonomously.}
       \label{fig:intuit}
    \end{figure}

Contact-based characterization techniques such as contact profilometry, four-point probes, and nanoindentation, among many others, are valuable tools in quantifying materials' surface \cite{piegari1985thin, brown1991describing, zhu2007nanoindentation, minor2006new, custance2009atomic} and electrical properties \cite{su2024intelligent, ebbesen1996electrical, wang2020probing, bash2022accelerated, chandra2017open, shimanovich2014four, sun2013conductivity}. By integrating deep learning and autonomous robotics into these methods, we can improve the reliability and quality of measurements \cite{soori2023artificial, chen2020deep, hippalgaonkar2023knowledge}, relieving researchers from the burden of constantly monitoring experiments to ensure optimal performance. However, integrating autonomy into these contact-based methods of characterization faces the challenges of reliably predicting high-precision contact positions \cite{nejat2003high, unal2015design, li2019survey}, establishing high-throughput feedback control \cite{leveziel2022migribot, kuwata2009real}, and collecting large labeled training datasets \cite{sunderhauf2018limits, thompson2020computational}. Deep learning-controlled robotic measurement of material and molecular properties has been widely implemented across a range of optical characterization techniques \cite{rapp2024self, wang2022data, siemenn2024using, omidvar2024accelerated, zhao2023robotic, azizi2024autonomous, mahjour2023rapid}, due to their non-contact nature, which simplifies mechanical complexity and increases data acquisition throughput compared to contact-based methods. For example, Su et al. \cite{su2024intelligent} propose a robotic non-contact atomic force microscopy (nc-AFM) probe with positional control driven by the Faster Region-based convolutional neural network (Faster R-CNN) \cite{siradjuddin2021faster}, which detects a general spatial bounding box around a target molecule for fast real-time positioning. However, the general bounding box is insufficient for orienting the pose of the robot to pixel-precise positions, and the model may require collecting additional labeled image data for fine-tuning \cite{su2024intelligent}. Instead, the implementation of a self-supervised approach designed specifically for spatial positioning tasks has the potential to address these challenges, resulting in high-precision predictive robotic control that more closely emulates the intuition of a human domain expert.

Here, we propose the design of a self-supervised and spatially differentiable convolutional neural network (SDCNN) for optimal pose prediction of contact-based robotic characterization systems and a reliable graph-based path planner to maximize measurement throughput across predicted poses. We utilize this SDCNN and path planner to precisely and autonomously control a 4-degree-of-freedom (4DOF) robot with a four-point probe end effector to make optimal contact with each film and measure photoconductivity without going out of bounds, \textit{i.e.}, a valid robot contact pose. Each film is drop-casted using only 4 $\upmu$L of chemical precursor, which allows us to maximize the number of combinatorial perovskite compositions explored but produces small-area films that are difficult to characterize using contact-based approaches. Hence, implementing spatial differentiability into a CNN enables the computer vision-segmented films to be used as shape priors in the loss function for the refinement of predictions directly in image space, transforming an unsupervised learning problem into a self-supervised one. We demonstrate the general-use nature of the proposed SDCNN on two different characterization tasks of perovskite semiconductors: (1) surface profilometry and (2) photoconductivity. The robot pose prediction performance is then evaluated across four metrics: (1) positional accuracy, (2) rotational accuracy, (3) valid pose generation, and (4) inference time. Our model's performance across these metrics is compared to seven other CNN models with either conventional loss functions or robust loss functions from literature, such as Wing \cite{wing}, Reverse Huber \cite{berhu}, and Barron \cite{barron}, that are designed for spatial tasks. Our approach achieves a 20.0\% improvement in the generation of valid poses, a 1.5\% improvement in positional accuracy, and equivalent rotational accuracy and inference time compared to the robust loss functions from literature \cite{wing, berhu, barron}. We develop a path planner to maximize the throughput of the autonomous characterization procedure by reliably generating distance-minimizing robot travel routes across predicted contact poses using stochastic noise. Generated path lengths are compared across four other planners from literature, such as A*, Christofides, Dijkstra, and a genetic algorithm. Our approach achieves a 5.0\% decrease in median path length generation with a 529.1\% tighter variance. These performance improvements of the SDCNN and planner approach enable the practical viability of our autonomous robot execution of contact-based measurements. We demonstrate this by autonomously characterizing $3,025$ photoconductivity curves of drop-casted perovskite semiconductors within 24 hours using the SDCNN-controlled 4DOF robotic system, achieving high throughputs of over 125 measurements per hour. These experimentally characterized photoconductivity values are then spatially mapped back onto each semiconductor film, resolving compositional trends of photoconductivity and identifying regions of non-uniformity where potential defects or degradation may exist. By achieving high-precision robotic spatial control for the characterization of semiconductors through self-supervised deep learning models and reliable path planning, we unlock new potential for integrating autonomous robotics into the semiconductor development and discovery pipeline, ultimately improving the reliability and quality of measurements at high throughputs without human supervision.

\section*{Results}

\subsection*{Autonomous robotic control through vision and deep learning}

\begin{figure}[h!p]
    \begin{center}
    \includegraphics[width=0.88\columnwidth]{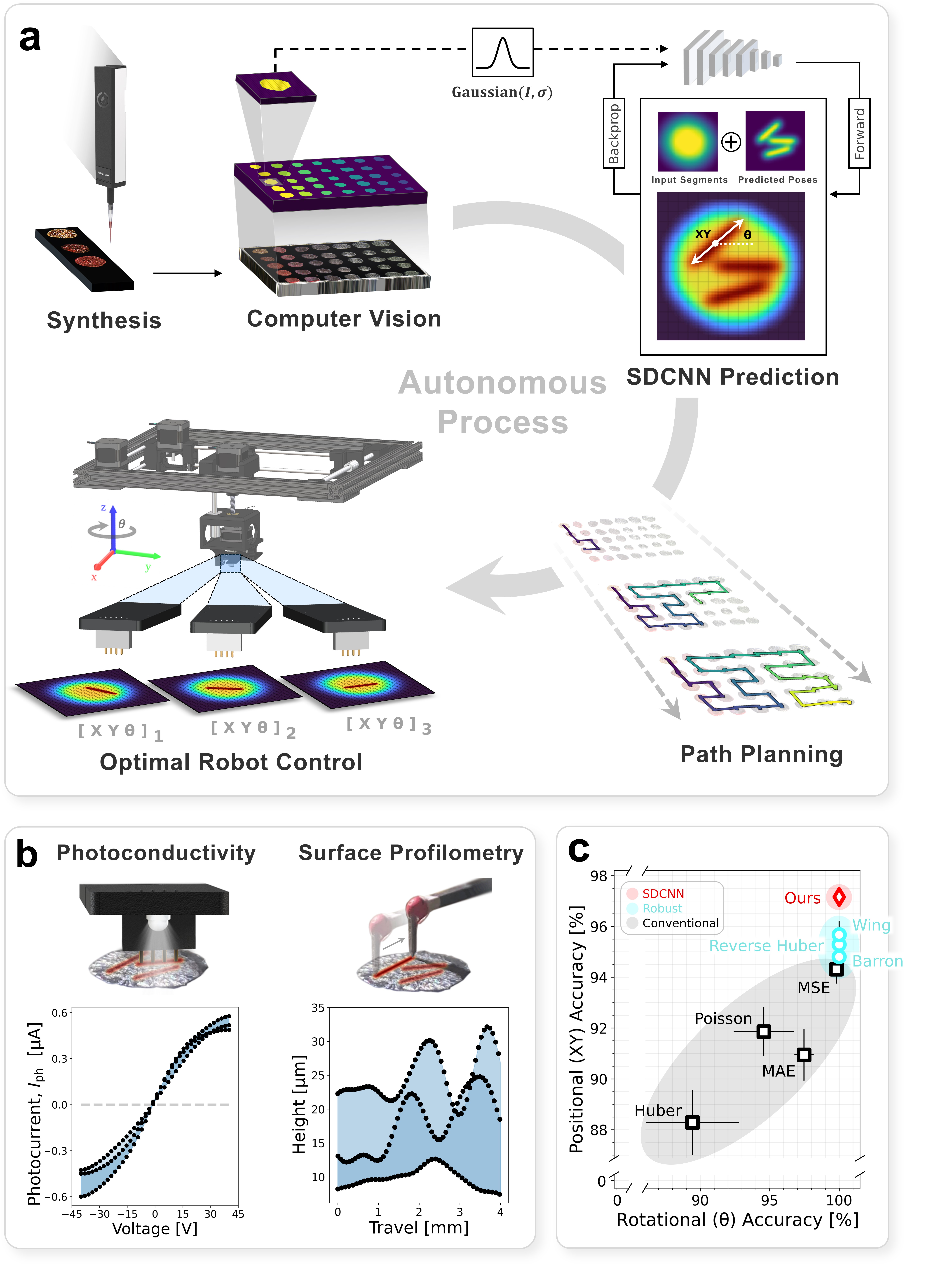}
    \end{center}
       \caption{Self-supervised robotic approach for autonomous contact-based characterization of semiconductors. \textbf{a}, Synthesized and annealed drop-casted semiconductor films are fed into the autonomous process, which begins with computer vision segmentation of the semiconductor films, then the prediction of optimal robot poses using a spatially differentiable CNN (SDCNN), distance-minimizing path planning, and finally robotic control with subsequent measurement. \textbf{b}, Photoconductivity and surface profilometry use-cases of the robotic contact-based characterization process. \textbf{c}, Average positional and rotational accuracy across $3,500$ robot pose predictions with two standard deviation error for our SDCNN method, compared against seven other models.}
       \label{fig:workflow}
    \end{figure}

Through optimal pose prediction from computer vision inputs, we autonomously drive a 4DOF robot with an illuminating four-point probe end effector to characterize the photoconductive properties of semiconductors spatially. Figure \ref{fig:workflow}a illustrates this autonomous control pipeline. Firstly, semiconductor films are drop-casted offline from the autonomous feedback loop using an OpenTrons volumetric pipetter \cite{mcgee2014screening}. Then, an on-board camera takes an image of the semiconductors, $\Phi_\textrm{cam}$, which is then rectified from the camera reference frame to the robot reference frame using a series of calibration matrices, $K$: 
\begin{equation}
    \Phi_\mathrm{robot} = K_\mathrm{img \rightarrow robot} \left( K_\mathrm{cam \rightarrow img} \Phi_\mathrm{cam} \right) \; ,
\end{equation}
where $K_\mathrm{cam \rightarrow img}$ rectifies $\Phi_\textrm{cam}$ from the camera to the image reference frame and then $K_\mathrm{img \rightarrow robot}$ rectifies $\Phi_\textrm{img}$ from the image to the robot reference frame. Once, in the same coordinate frame as the robot, the Fast Segment Anything Model (FastSAM) \cite{zhao2023fastsegment} is used to quickly find the edges of each drop-casted film, creating image segments, $I$. Next, optimal poses are predicted directly onto the image segments using the proposed SDCNN model, an 8-layer CNN with an objective-based spatially differentiable loss function derived using domain information. The optimality of each pose is determined using $I$ as a prior in image space, which can be back-propagated into the network as a loss due to spatial differentiability. We develop a noisy Dijkstra's planner to find a distance-minimizing path plan for traversing all predicted poses across all drop-casted films. Finally, the robot is controlled to each SDCNN-predicted pose to characterize the properties of the semiconductor at that location in space. The average positional and rotational accuracies of predicted poses by the SDCNN are shown to be higher than predictions by CNNs using existing loss functions (Fig. \ref{fig:workflow}c), with improvements between 1.5\% and 8.9\%. Designing the SDCNN to predict poses with high positional accuracy and rotational uniqueness enables comprehensive spatial mapping of important measured semiconductor properties in the fewest number of poses.

The spatial mapping capabilities of this autonomous workflow can be generalized to function with different contact-based end effectors. Figure \ref{fig:workflow}b highlights two characterization use cases: photoconductivity and surface profilometry. Photoconductivity is measured at each predicted pose by taking the difference between illuminated and dark current-voltage curves (Fig. \ref{fig:workflow}b, left). The blue-shaded regions highlight the spread of the photoconductive properties across the area of a drop-casted methylammonium lead iodide (MAPbI$_3$) perovskite film for a set of $k=3$ poses, predicted by the SDCNN to maximize the unique spatial area measured. These results indicate that photoconductivity is generally uniform across the area of this particular film. Conversely, the thickness of the film varies largely by tens of micrometers, measured using surface profilometry at the same spatially predicted contact poses (Fig. \ref{fig:workflow}b, right). Utilizing different end effectors with the same driving SDCNN demonstrates the model's generalizable nature for spatially resolving important properties of semiconductor materials in an autonomous fashion.

\subsection*{Spatial differentiability for optimal robot pose prediction}

\begin{figure}[h!]
    \begin{center}
    \includegraphics[width=1.\columnwidth]{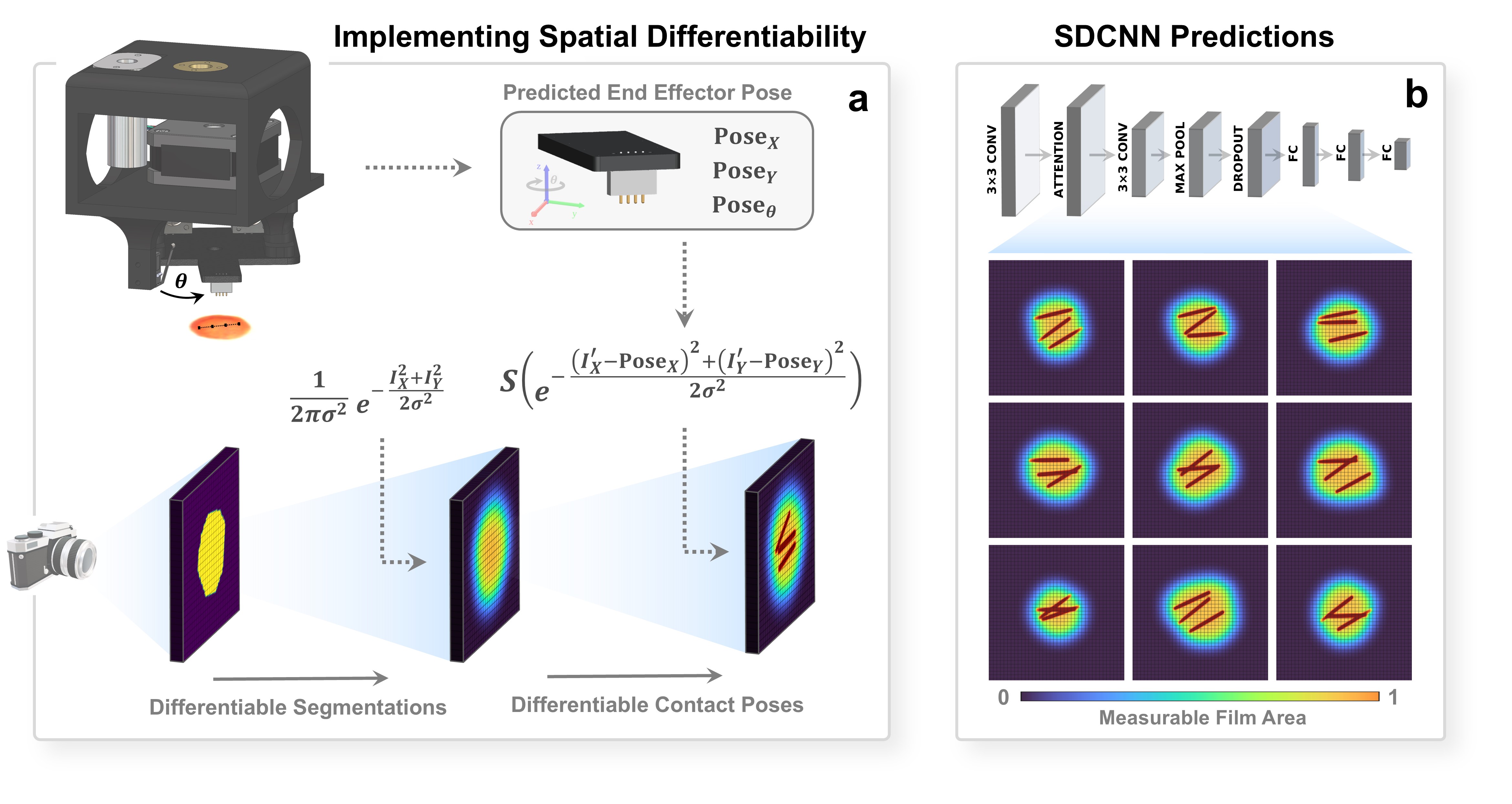}
    \end{center}
       \caption{Implementing CNN spatial differentiability for self-supervised robot pose optimization. \textbf{a}, Image segment pixels ($I_X, I_Y$) are passed through a Gaussian filter to maintain the differentiability of the edges. Predicted pose pixels, $(\mathrm{Pose}_X, \mathrm{Pose}_Y)$, from the SDCNN are superimposed onto the differentiable segment pixels, ($I'_X, I'_Y$), using another Gaussian filter and sigmoid function, $S(\cdot)$, to perform direct computation and back-propagation in image space. \textbf{b}, SDCNN architecture and differential predicted poses, $(\mathrm{Pose}'_X, \mathrm{Pose}'_Y)$, composed onto the image segment pixels, ($I'_X, I'_Y$). The SDCNN is trained to predict poses onto the measurable regions of the films, sufficiently distanced from its edges.}
       \label{fig:construct}
    \end{figure}

Spatial differentiation ensures the loss function is differentiable across image pixels, enabling back-propagation of image-based computations through SDCNN neurons. Here, we use spatial differentiability to convert unsupervised learning into self-supervised learning by creating a pixel-based loss function using domain information that accepts shape priors as inputs to refine a prediction. We aim to have the SDCNN predict a set of $k$ valid robot poses that will maximize the number of pixels making unique contact with the spatial area of each film. The image segments, $I$, and predicted poses, $\{ \mathrm{Pose}_i \}_{i=1}^{k}$, are the shape priors to the loss function. This method is valid for convex shapes in general and does not assume a perfect circle (Fig. \ref{sfig:shapes}). These shapes are first passed through Gaussian filters to ensure they are smooth and differentiable before being passed to the loss function, generating, $I'$ and $\{ \mathrm{Pose}'_i \}_{i=1}^{k}$ (Fig. \ref{fig:construct}a). A predicted pose is considered a valid contact if all its pixels fall within the measurable area (non-zero valued region shown in Fig. \ref{fig:construct}b) of the differentiable segment $I'$, which can be tuned by modifying the standard deviation of the Gaussian filter, $\sigma$ (Fig. \ref{fig:construct}a). If any predicted contact pose pixels fall within the background area (zero-valued region), the pose is considered invalid. The spatially differentiable loss function consists of a tunable weighted sum of two optimization objectives derived from domain information commonly followed when characterizing materials using contact-based methods: (1) maximize the coverage of all poses in the measurable area without overlap and (2) maximize the spatial and rotational uniqueness of all poses. By embedding this domain information as an objective-based loss function, the network learns to predict poses more aligned with the measurement intuition of a domain expert, improving the autonomous decision-making capabilities of the robot:
\begin{equation}
\begin{split}
\label{eq:loss}
\mathrm{loss}\left( I', \{ \mathrm{Pose}'_i \}_{i=1}^{k} \right) = &
- \underbrace{
\omega_1 \sum_{i=1}^{k} \left(  I'_{XY} \circ \{ \mathrm{Pose}'_{XY_i} \}_{i=1}^{k} \right)
}_{\text{Maximize coverage within a segment}}
- \underbrace{
\omega_2 \operatorname{Var}\left( \{ \mathrm{Pose}'_{\theta_i} \}_{i=1}^{k} \right) \vphantom{\sum_{i=1}^{k}}
}_{\text{Maximize angular variance}}
\\[1ex]
& \qquad \qquad
\text{subject to} \quad \mathrm{Pose}'_i \cap \mathrm{Pose}'_j = \emptyset, \quad \forall\, i \neq j ,
\end{split}
\end{equation}
where $\omega_1$ and $\omega_2$ are weights set to $\omega_1=\omega_2$ and $I'_{XY} \circ \{ \mathrm{Pose}'_{XY_i} \}_{i=1}^{k}$ is the composition of the differentiable segment pixels onto the differentiable pose pixels. $\mathrm{Pose}'_{XY_i}$ and $\mathrm{Pose}'_{\theta_i}$ are the $XY$-pixel coordinates and the yaw-rotation angles, respectively, for $i\in \{1,..,k\}$ unique differentiable poses. The objectives are negated to form a loss and are subject to the constraint of reducing the number of overlapping pixels between any two unique predicted poses.

\begin{figure}[p!h]
    \begin{center}
    \includegraphics[width=0.82\columnwidth]{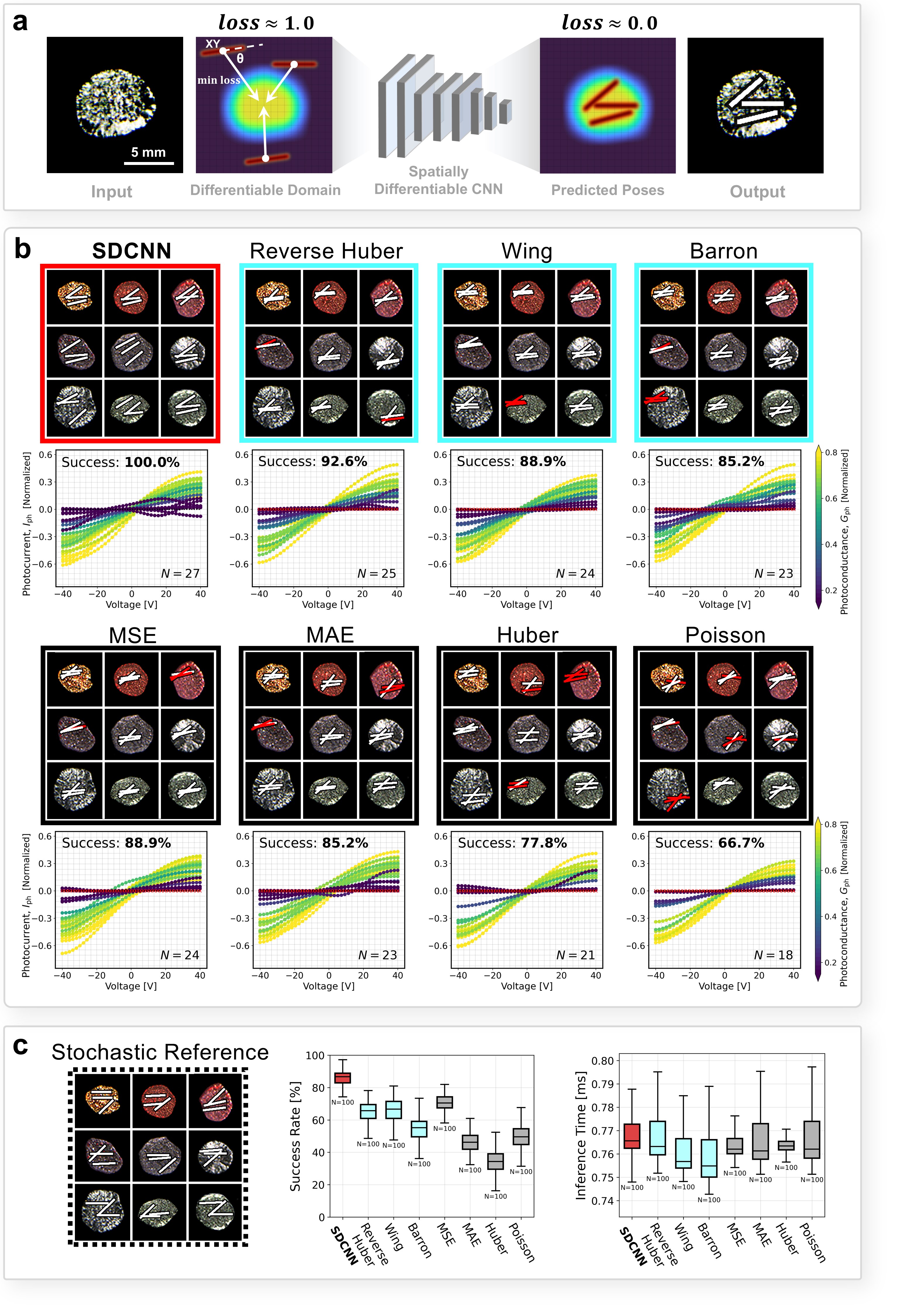}
    \end{center}
       \caption{Pose prediction performance across CNN models on drop-casted semiconductor films with varying geometries. \textbf{a}, Loss minimization procedure in image space using our spatially differentiable CNN (SDCNN). \textbf{b}, Predicted poses for eight different CNN models on a subset of nine experimentally synthesized semiconductor films for $k=3$ poses per film. Predicted valid contact poses are shown in white, and invalid poses are shown in red. The photocurrent curve at each valid pose is measured experimentally using the robotic system. \textbf{c}, Left: stochastic process predictions. Middle: prediction success rate on the full set of 35 experimentally synthesized semiconductors across 100 unique trials. Right: CNN inference time on the pose prediction task for the full set of films across 100 unique trials, run on an NVIDIA Tesla V100 GPU.}
       \label{fig:cnn}
    \end{figure}

 
Figure \ref{fig:cnn}a illustrates this loss minimization procedure in image space. Before training, the network has a high loss since it has not learned how to use the shape priors, resulting in placing poses randomly. After training on computer vision-segmented semiconductor film borders, the network learns to place poses in unique positions within the measurable regions without overlap using the segments as priors to the objective-based loss function. Figure \ref{fig:cnn}b compares the performance on the valid pose prediction task of our SDCNN against seven other CNN models that use existing loss functions: recent loss functions from literature designed for robust and spatial tasks (Reverse Huber \cite{berhu}, Wing \cite{wing}, and Barron \cite{barron}) and conventional loss functions (mean squared error (MSE), mean absolute error (MAE), Huber, and Poisson negative log-likelihood). Corresponding photoconductivity curves are experimentally measured on the perovskite film using the 4DOF robotic system for each valid predicted pose. The models using existing loss functions tend to cluster predicted poses tightly together, often resulting in close or partial overlaps. In contrast, our approach encourages predictions to be more spread apart, effectively utilizing the full segment prior, while staying within the measurable area. Furthermore, in addition to tight clustering, as shown in Fig. \ref{fig:cnn}b, the existing models also have lower positional accuracy (Fig. \ref{fig:workflow}c), resulting in fewer valid poses being successfully generated for this subset of nine perovskite films, compared to our spatially differentiable model. 

In Fig. \ref{fig:cnn}c (middle), we expand this analysis to the full set of 35 films and perform 100 replicate trials of valid pose generation. Across these 100 trials, we demonstrate that our SDCNN model achieves improvements of 20.0\% and 16.2\% in the median success rate of generating valid poses over robust loss functions and conventional loss functions, respectively. Given these performance improvements, the inference time of our loss function is negligible compared to the hardware response time, increasing by only 2.4 ns, relative to the slowest tested loss function in our evaluation, Reverse Huber (Fig. \ref{fig:cnn}c, right). Hence, utilizing spatial differentiability in a neural network for pose prediction tasks improves the reliability of autonomous robotics by successfully generating valid predictions without sacrificing compute speed while also reducing the data labeling burden through self-supervision.

\subsection*{Reliable graph-based robot path planning}

A reliable path planner consistently generates optimal path plans with low variance across multiple executions of a given task. Optimality is defined here as minimizing the total travel distance of the robotic probe to execute a plan. Hence, a planner that can reliably generate time-efficient plans for the robotic probe increases the overall system throughput across autonomous experimental cycles. The planning task at hand is to move from a fixed start node and visit all other nodes in the graph exactly one time while accumulating the shortest travel distance. Each pose predicted by the SDCNN represents a node in a graph, where all nodes together form the full graph. This problem is akin to the classic Traveling Salesman Problem (TSP), which is an NP-hard optimization problem \cite{junger1995traveling}. Our application differs from the traditional TSP such that we relax the constraint of returning back to the start node -- we define this as an Open Loop Traveling Salesman Problem (OTSP) \cite{chieng2014performance}. 

\begin{figure}[h!p]
    \begin{center}
    \includegraphics[width=1\columnwidth]{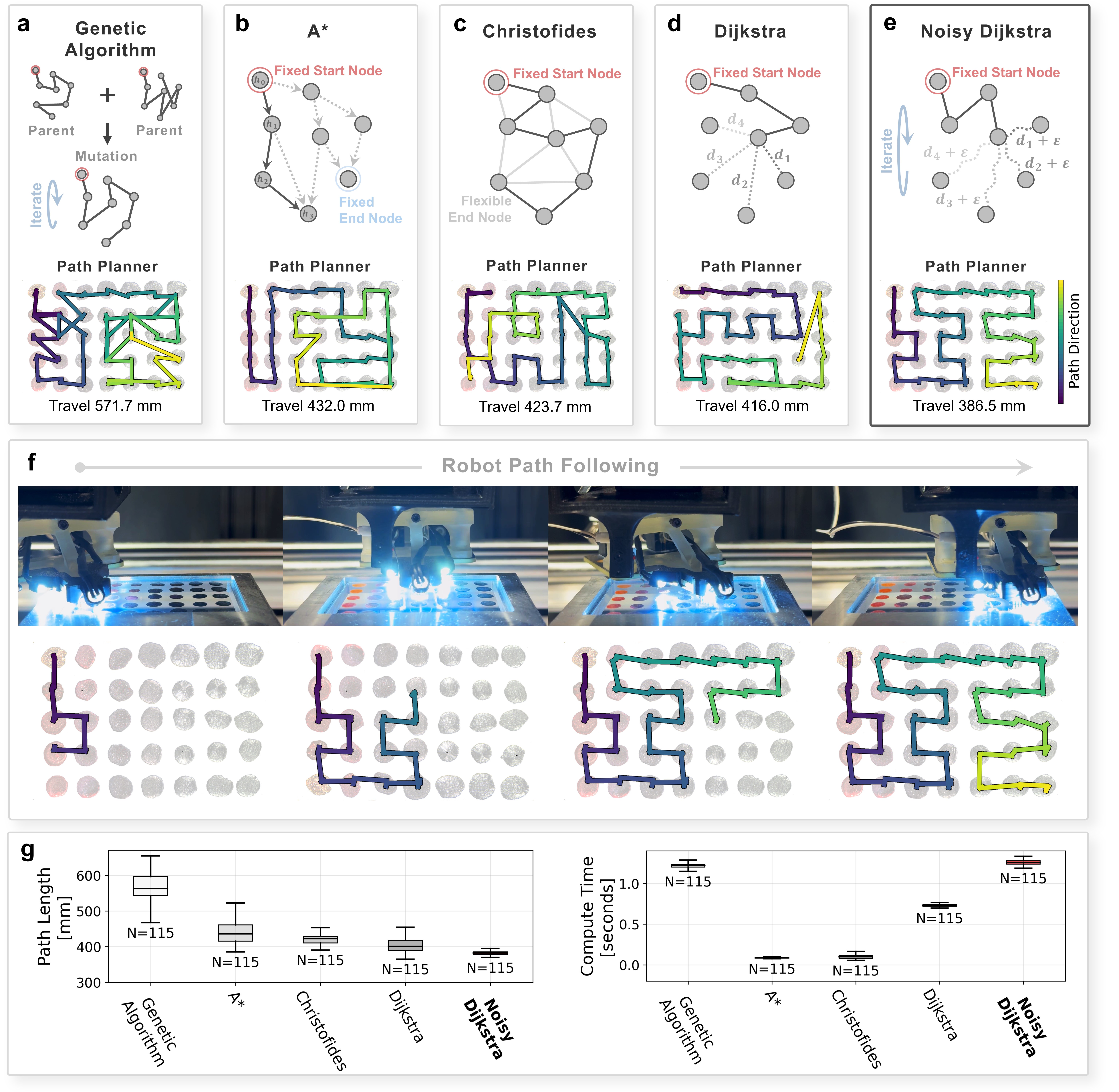}
    \end{center}
       \caption{Distance-minimizing path planning performance across graph-based algorithms for autonomous contact-based robotics. \textbf{a--e}, Demonstration of the planning mechanics for five path-finding algorithms to solve an Open Loop Traveling Salesman Problem (OTSP). A path plan generated by each of these methods is illustrated by the color map, indicating the path direction for a graph of nodes, each node representing a robot contact pose predicted by the spatially differentiable CNN (SDCNN) on an array of drop-casted semiconductor films. \textbf{f}, The robot following a path generated by our proposed noisy Dijkstra path planner. \textbf{g}, Left: total path lengths of generated routes across $115$ independent trials, each computed on a unique graph of $k=3$ SDCNN-predicted robot contact poses for an array of 35 drop-casted semiconductor films. Right: compute time to generate a path plan across $115$ independent trials, run on an NVIDIA GeForce RTX 4090 GPU.}
       \label{fig:plan}
    \end{figure}

Several methods exist for generating global approximations of the TSP and OTSP, including genetic algorithms (GA) \cite{potvin1996genetic}, A* \cite{Hart1968}, Christofides algorithm \cite{christophides1976worst}, and Dijkstra's algorithm \cite{dijkstra1959note}. GAs introduce randomness into the optimization of distance-minimizing graphs by iteratively combining and mutating parent graphs to create new offspring graphs. The A* algorithm uses a distance heuristic, $h$, estimating the cost from the current node to the target node to guide the search and find the shortest path. The Christofides algorithm uses the triangle inequality to connect all nodes while minimizing added distance, ensuring the total path distance does not exceed $1.5$x the optimal path length. Dijkstra's algorithm calculates the distances, $d$, between a current node and all other unvisited nodes to build a graph that connects all nodes together with the shortest distances. However, for our task, $k$ predicted contact poses, \textit{i.e.} nodes, are clustered within the area of each semiconductor film, resulting in these existing methods having trouble finding efficient paths. Redundant overlaps and loops create inefficiencies that increase the total travel distance and path variance across multiple executions (Fig. \ref{fig:plan}a--d). To overcome this challenge, we introduce a noisy Dijkstra's algorithm that blends the distance-minimizing graph-building procedure from Dijkstra's algorithm with the stochastic process of the GA by adding noise, $\varepsilon$, to each distance value to minimize the total travel distance over several generations, generating more reliable path plans without overlaps or loops (Fig. \ref{fig:plan}e--f):
\begin{equation}
\begin{split}
    \min_{|P|} \sum_{(i,j) \in P} \left(d_{ij} + \varepsilon_{ij}\right),\\
    \text{where } \varepsilon_{ij} \sim \mathcal{U}\left(-\alpha \, d_{ij},\ \alpha \, d_{ij}\right).
\end{split}
\end{equation}
where $|P|$ is the total length of a generated path, $P$, in the graph. $d_{ij}$ is the length of each edge in the graph and $\varepsilon_{ij}$ is the added stochastic noise to each edge length, where each $\varepsilon_{ij}$ is independently sampled for each edge from a uniform distribution within $\pm \alpha$ of the edge length.

In Fig. \ref{fig:plan}g, we compare the generated total path lengths and GPU compute time for all five methods across 115 uniquely SDCNN-predicted graphs for arrays of 35 drop-casted gradients of semiconductor films. Through stochastically minimizing total path length with the addition of edge length noise, our proposed noisy Dijkstra approach reliably generates direct and non-overlapping distance-minimizing paths for the OTSP task. Across 115 graphs, noisy Dijkstra's algorithm achieves a median total path length of $381.7$ mm -- a $5.0$\% improvement over standard Dijkstra's algorithm -- with a variance of $34.3$ mm$^2$ -- a $529.1$\% improvement over the Christofides algorithm, which are the highest performing methods tested from literature for these metrics. Figure \ref{fig:plan}f illustrates the robot following an efficient path plan generated by this noisy Dijkstra approach across the graph of SDCNN-predicted contact poses. Furthermore, for the same number of generations, both noisy Dijkstra and the GA achieve comparable GPU compute times of $1.22$ seconds per path and $1.26$ seconds per path, respectively, despite noisy Dijkstra improving median path length by $47.3$\% and variance by $4,249.3$\% relative to the GA. These results demonstrate that our noisy Dijkstra's algorithm is both an efficient and reliable path-planning approach for autonomous contact-based robotic measurement, outperforming existing methods in generating shorter paths with tighter variance while maintaining comparable computational speeds.

\subsection*{Robotic spatial mapping of photoconductivity with autonomy}


Coupling our developed self-supervised SDCNN for pose prediction with the noisy Dijkstra path planner, this 4DOF robot -- using an illuminating four-point probe end effector -- enables high-throughput photoconductivity characterization by emulating expert intuition in selecting optimal contact points and generating efficient path plans without human intervention. This implementation of autonomy enables the robot to make domain-informed decisions about where to contact each material with its end effector and in what order to make these contacts, ultimately improving the reliability, quality, and throughput of the measurements. As many semiconducting materials such as solar cells and light-emitting diodes (LEDs) can now be rapidly synthesized and manufactured with self-driving laboratories \cite{macleod2020self, langner2020beyond, bag2016rapid, luo2021high, son2012discovery}, it is critical to ensure accompanying measurement techniques also increase in throughput while upholding human-level quality and reliability. 

\begin{figure}[p!h]
    \begin{center}
    \includegraphics[width=1.\columnwidth]{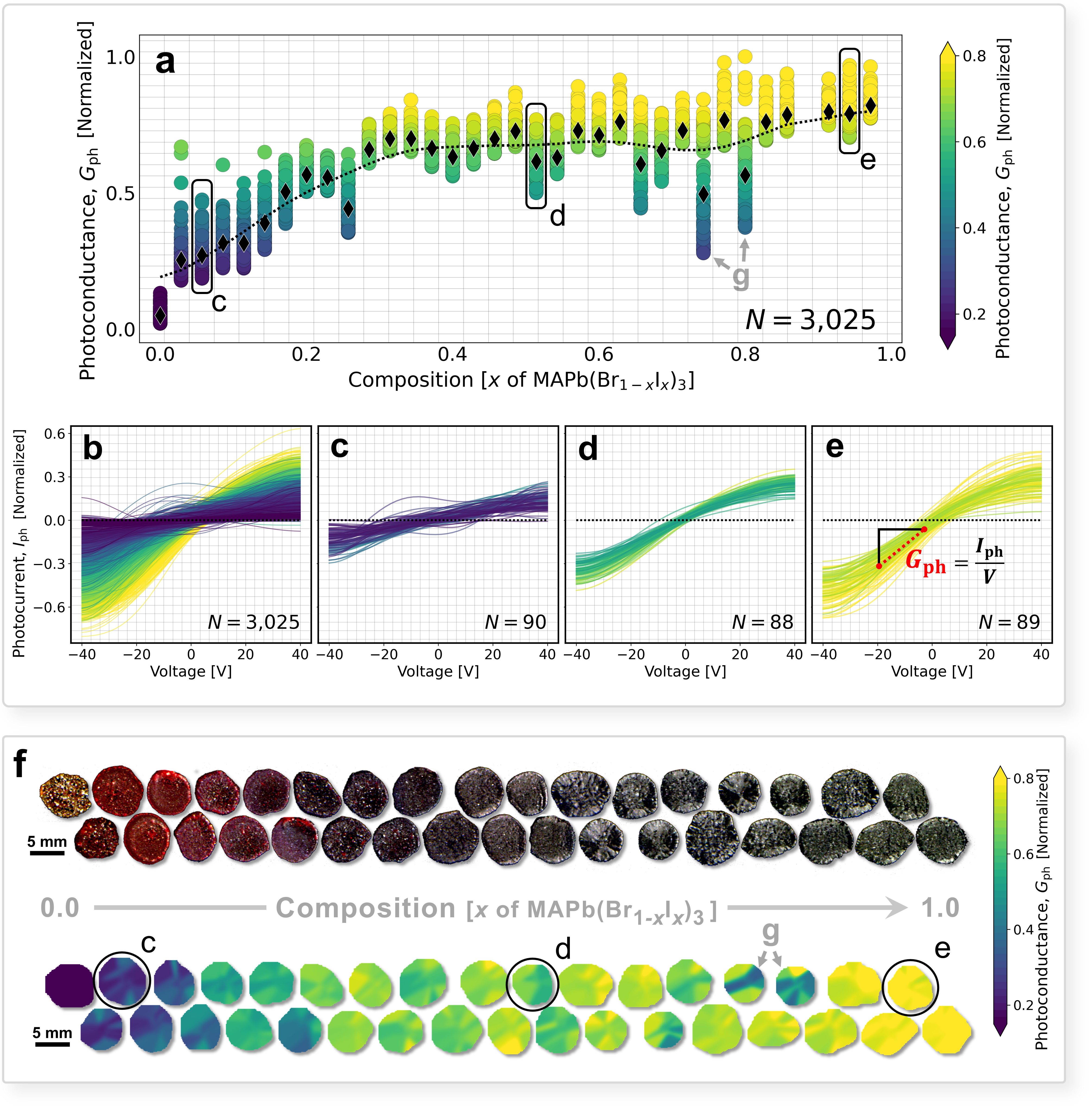}
    \end{center}
    \caption{Photoconductivity of perovskite semiconductors characterized using a self-supervised, autonomous robotic system. \textbf{a}, Photoconductance, $G_\mathrm{ph}$, measured at $3,025$ unique poses, predicted by the spatially differentiable CNN (SDCNN), across a gradient of drop-casted methylammonium lead bromide (MAPbBr$_3$) to methylammonium lead iodide (MAPbI$_3$) mixed-halide perovskite semiconductor films, MAPb(Br$_{1-x}$I$_x$)$_3$. \textbf{b}, All measured photocurrent, $I_\mathrm{ph}$, curves with colormap corresponding to $G_\mathrm{ph}$ (slope of the photocurrent-voltage curve). Measured $I_\mathrm{ph}$ curves are displayed for the following perovskite compositions: \textbf{c}, MAPb(Br$_{0.94}$I$_{0.06}$)$_3$, \textbf{d}, MAPb(Br$_{0.47}$I$_{0.53}$)$_3$, and \textbf{e}, MAPb(Br$_{0.03}$I$_{0.97}$)$_3$. \textbf{f}, Top: images of synthesized MAPb(Br$_{1-x}$I$_x$)$_3$ films. Bottom: spatially mapped $G_\mathrm{ph}$ with Gaussian interpolation, collected using the autonomous robotic system. \textbf{g}, Perovskite films with highly non-uniform spatially-resolved $G_\mathrm{ph}$.}
    \label{fig:photocurrent}
\end{figure}

To demonstrate measurement performance, we run the developed robotic system autonomously for 24 hours in continuous operation without human intervention to measure the photoconductivity of drop-casted gradients of methylammonium lead bromide (MAPbBr$_3$) to methylammonium lead iodide (MAPbI$_3$) mixed-halide perovskite films, MAPb(Br$_{1-x}$I$_x$)$_3$. Over the course of this 24-hour campaign, $3,025$ unique photoconductivity measurements are taken by the SDCNN-controlled robotic system, resulting in a characterization throughput of over 125 measurements per hour. At each SDCNN-predicted pose, the photocurrent, $I_\mathrm{ph}$, is measured as a function of voltage, $V$, by taking the difference between the illuminated and dark current-voltage curves. Then, photoconductance, $G_\mathrm{ph}$, can be characterized at each pose by computing the slope of each photocurrent-voltage curve: 
\begin{equation} 
\label{eq:photo}
G_\mathrm{ph} = \frac{I_\mathrm{ph}}{V} = \frac{I_\mathrm{light} - I_\mathrm{dark}}{V} \; , 
\end{equation} 
where $I_\mathrm{light}$ is the current measured under illumination and $I_\mathrm{dark}$ is the current measured in the dark. Figure \ref{fig:photocurrent}a shows all characterized values of $G_\mathrm{ph}$ as a scatter plot across the gradient of drop-casted perovskites. The distribution of $G_\mathrm{ph}$ along the $y$-axis for each composition illustrates the spatial variance of photoconductance across the area of each film with median values shown as black diamonds. An increasing trend in $G_\mathrm{ph}$ is observed (dashed curve in Fig. \ref{fig:photocurrent}a) as the composition shifts from MAPbBr$_3$ to MAPbI$_3$ under broad-band white light illumination. This is consistent with the decreasing bandgaps of the corresponding perovskite compositions from 2.3 to 1.6 eV \cite{jang2015reversible}, assuming similar thicknesses between films. These results confirm that the robotic system not only achieves a high measurement throughput but also reliably captures the expected trends in photoconductivity across composition gradients.

Figure \ref{fig:photocurrent}b displays the full experimentally measured photocurrent-voltage curves, measured at each predicted pose. The spatially characterized $I_\mathrm{ph}$ curves are highlighted for three perovskite films within the MAPb(Br$_{1-x}$I$_x$)$_3$ gradient: bromine-rich (Fig. \ref{fig:photocurrent}c), mixed (Fig. \ref{fig:photocurrent}d), and iodine-rich (Fig. \ref{fig:photocurrent}e). Data collected across several spatially distinct contact points for a given film allows us to generate a spatial map of experimental $G_\mathrm{ph}$ values using a simple Gaussian interpolation (Fig. \ref{fig:photocurrent}f, bottom). Detailed spatial mapping is critical for identifying defects or non-uniformities in material synthesis, which can significantly affect device performance. For example, based on the trend observed in Fig. \ref{fig:photocurrent}a, we expect iodine-rich films to have higher $G_\mathrm{ph}$ values across the area of the film area with some expected spatial variation. However, in certain instances of iodine-rich films (Fig. \ref{fig:photocurrent}g), we observe regions of unexpectedly low $G_\mathrm{ph}$, likely induced by early degradation or pinhole defects in the film. Hence, by rapidly resolving spatial properties and accelerating the detection of critical non-uniformities, the developed autonomous and high-throughput robotic characterization system capabilities ensure that measurement techniques keep pace with the rapid synthesis of semiconductors in self-driving laboratories while upholding human-level quality and reliability needed to identify performance-limiting defects \cite{WIEGHOLD2017526, kunze2022}.


\section*{Discussion}

In this paper, we develop an approach to enable autonomous and domain-informed characterization of semiconductor materials using contact-based robotics to improve measurement reliability, quality, and throughput. Firstly, we develop a self-supervised spatially differentiable convolutional neural network (SDCNN) for reliable and accurate pose prediction of contact-based autonomous robotics that optimizes predictions based on domain information to more closely emulate a human expert's measurement intuition. We demonstrate that this self-supervised SDCNN improves the generation of valid robot contact poses by up to $20.0$\% with improved positioning and spatial coverage compared to existing supervised methods \cite{wing, berhu, barron}. Secondly, we develop a noisy Dijkstra's algorithm graph-based path planner that combines distance-minimizing graph-building approaches with stochastic genetic approaches to generate efficient measurement routes that reliably minimize the total robot travel distance. We demonstrate that this noisy Dijkstra planner generates efficient measurement routes with few to no overlaps or loops while decreasing planning variance across experiments by over 6x compared to existing graph-based methods \cite{guofei9987_scikitopt, Hart1968, christophides1976worst, dijkstra1959note}. Through the coupling of these developed methods, we perform autonomous characterization of over $3,000$ photoconductivity curves from drop-casted mixed-halide MAPb(Br$_{1-x}$I$_x$)$_3$ perovskite semiconductor films in 24 hours using a 4DOF robotic probe, achieving high throughputs of over 125 measurements per hour without human intervention. By using the domain-informed SDCNN to predict a minimal set of robot contact poses that adequately cover each uniquely shaped film, we quickly obtain a high-resolution spatial map of experimentally measured material properties. Rapidly resolving spatial properties offers early insight into the formation of defects and degradation, providing a tool to quickly and autonomously detect performance issues in semiconductor fabrication lines and self-driving laboratories \cite{WIEGHOLD2017526, kunze2022}. Altogether, this approach advances the methods of high-precision and efficient robotic spatial control in material characterization, taking a step toward fully autonomous integration while ensuring high-throughput and reliable measurements without human supervision.


Although the proposed robotic system is designed for autonomous operation from input image to experimental measurement, its calibration and swapping of end effectors are currently limited to manual operation (Fig. \ref{sfig:calibration}). This manual calibration process is often tedious and relies heavily on the skill and precision of the user, leading to varying results and potential inconsistencies in experimental outcomes \cite{nejat2003high}. To minimize this source of variance and enhance the reliability of the robotic system, automated calibration techniques could be employed in future developments. For instance, automated bed leveling and height mapping are widely used in current 3D printing systems to ensure reliable performance between prints or tool changes \cite{hofbauer2023automatic}, utilizing sensors and control algorithms to detect and compensate for misalignments. Integrating similar methods, such as vision-based calibration systems \cite{enebuse2021comparative}, or machine learning for calibration tasks \cite{li2021data, kong2022precision}, could further enhance precision and adaptability. These advancements would reduce dependency on user expertise, improve overall efficiency, and broaden the system's applicability to quantify a wider range of significant material properties.

Moreover, the number of predicted robot poses by the SDCNN is currently limited to the output vector length specified during the training procedure. This means that changing the number of output poses requires retraining the entire model, which can be time-consuming. To address this limitation, future developments may involve incorporating conditional rules through mixtures of experts (MoE) \cite{zhu2022uniperceivermoelearningsparsegeneralist} or transformer-based architectures \cite{NIPS2017_3f5ee243, devlin2018bert}, which facilitate the generation of a conditional number of poses without requiring complete retraining. Incorporating dynamic neural networks \cite{han2021dynamic} and conditional variational autoencoders \cite{kingma2014semi} could further enhance the model's ability to adaptively generate poses tailored to specific experimental requirements. Additionally, integrating techniques from active learning \cite{baranes2013active} could enable the model to selectively update itself with new data, reducing the need for extensive retraining. These future works aim to extend the accessibility of the developed system to users without domain expertise in robotics while broadening the versatility of the applied deep learning models.

With the developed self-supervised SDCNN model and graph-based noisy Dijkstra path planner integrated into a 4DOF robotic system, we have demonstrated the implementation of reliable autonomy with minimal data overhead into the contact-based characterization of semiconductor films, measuring at high throughputs of over 125 photoconductivity curves per hour. Our proposed method improves the automated quantification of critical surface and electrical material properties, addressing prior challenges in achieving reliable and fast automation. This advancement in coupling deep learning and robotics for materials science takes a key step toward improving and accelerating the tools and methods for autonomous spatial characterization of critical semiconductor properties, in turn, galvanizing the pipeline of self-driving materials research.

\section*{Methods}

\subsection*{Perovskite material preparation}

To prepare the MAPb(Br$_{1-x}$I$_x$)$_3$ gradient of perovskite semiconductors, we use OpenTrons mixing and drop-casting of 0.6M MAPbI$_3$ and 0.6M MAPbBr$_3$ precursor solutions. The MAPbI$_3$ precursor is prepared using a 4:1 ratio of dimethylformamide (DMF, $\geq$99.8\%, Sigma-Aldrich) to dimethylsulfoxide (DMSO, $\geq$99.9\%, Sigma-Aldrich) solvent and then dissolving a 1:1 ratio of methylammonium iodide (MAI, $>$99.9\%, Greatcell Solar Materials) to lead iodide (PbI$_2$, 99.999\% trace metal basis, Sigma-Aldrich) solutes into the solvent mixture using a vortex mixer. The MAPbBr$_3$ precursor is prepared using a 4:1 ratio of DMF:DMSO solvent and then dissolving a 1:1 ratio of MAI to lead bromide (PbBr$_2$, 99.999\% trace metal basis, Sigma-Aldrich) solutes into the solvent mixture using a vortex mixer. Once the precursors are prepared, the OpenTrons pipettes gradated concentrations of each precursor into 35 smaller volume vials in serial. Then, mixing is induced for each of the 35 unique compositions by repeatedly aspirating and dispensing the fluid in the vials three times. Once mixed, 4 $\upmu$L of each solution is pipetted onto a glass slide that is pre-heated to 55$^\circ$C to form an array of individual films on the glass slide. Before heating, the glass slide was washed with isopropyl alcohol (IPA, $\geq$ 99.5\%, VWR). After drop-casting, the glass slide is transferred to a pre-heated hot plate and annealed at 150$^\circ$C for 20 minutes.

\subsection*{Robot and end effector design}

The 4DOF robot controlled by the SDCNN and used for the measurement of semiconductor properties is custom-built. The frame of the robot positioning system is built using 80/20 T-slotted aluminum extrusion rails. Three standard Nema 17 stepper motors and timing belts control the $XYZ$-positioning of the robot. A Bigtreetech Direct Octopus V1.1 control board with TMC2209 stepper motor drivers, each tuned for 0.75A of output current, are used to drive all Nema motors. The poses predicted by the SDCNN are converted to Marlin G-code, which is sent to the control board \textit{via} Python serial communication to execute motion commands. The motor that controls the $Z$-positioning is affixed to a 3D-printed chassis and uses a ball screw to drive the $Z$-positioning of the end effector mount. The last motor that controls the $\theta$-positioning (yaw) is a Nema 17 pancake motor affixed to a 3D-printed end effector mount. To this mount, an Ossila four-point probe head is affixed. We design an anti-cantilever attachment for the neck of the probe using rigid stereolithography 3D printing to mitigate $Z$-positioning drift. Attached to the head of the probe is a 3D-printed mount for three high-powered LEDs used to angle light to measure the $I_\mathrm{light}$ of the film in contact. The LED mount is designed to maximize uniform spread of light while minimizing shading, in turn, improving the consistency of measured $I_\mathrm{light}$.

\subsection*{Spatially differentiable loss function construction}

The spatially differentiable loss function accepts shapes as priors to refine predictions. However, to maintain differentiability within the spatial image domain, each shape's edge must be smoothed. The shapes that get passed through the loss function are the computer vision segmentations of the semiconductor films, $I$, and the set of $k$ poses predicted by the SDCNN, $\{ \mathrm{Pose}_i \}_{i=1}^{k}$. To maintain differentiability, the pixels of these shapes pass through a 2D Gaussian filter:
\begin{equation}
    I'_{XY} = \frac{1}{2 \pi \sigma^2} \exp{\left(-\frac{I^2_X + I^2_Y}{2 \sigma^2}\right)} \; ,
\end{equation}
where $\sigma$ is the standard deviation of the Gaussian and $I_{XY}$ are the non-differentiable $XY$-pixels of a segment. A composition is created by superimposing the predicted poses directly onto a differentiable segment:
\begin{equation}
    I'_{XY} \circ \mathrm{Pose}'_{XY} = S \left( \exp{\left(-\frac{(I'_X-\mathrm{Pose}_X)^2 + (I'_Y-\mathrm{Pose}_Y)^2}{2 \sigma^2}\right)} \right) \; ,
\end{equation}
where $S(\cdot)$ is the sigmoid function, used as a soft threshold for placement of the poses onto the segment, $I'_{XY}$ are the differentiable $XY$-pixels of a segment, and $\mathrm{Pose}_{XY}$ are the non-differentiable $XY$-pixels of a predicted pose. With this composition, differentiability within pixel space is ensured, and now all differentiable computations performed on the image can be backpropagated to the network weights during training. The loss function that we aim to minimize to train the network is expressed in Eq. \ref{eq:loss}.

\subsection*{Neural network architecture and training}

We evaluate the performance of 8 different CNN models on the pose prediction task, each with the same 8-layer network architecture but with a different loss function: SDCNN (ours), robust spatial loss functions from literature (Wing \cite{wing}, Reverse Huber \cite{berhu}, Barron \cite{barron}), and conventional loss functions (MSE, MAE, Poisson, Huber). Our 8-layer network consists of the following architecture: (1) $3\times3$ convolution with batch normalization, (2) spatial attention mechanism, (3) $3\times3$ convolution with batch normalization, (4) max pooling, (5) 50\% dropout, (6) fully connected (FC) with $1,200$ neurons, (7) FC with $600$ neurons, (8) FC with $(k * 3)$ neurons, for the (1) $X$-coordinate midpoint, (2) $Y$-coordinate midpoint, and (3) rotation angle, $\theta$, of $k$-number of poses: $\{(\mathrm{Pose}'_{X_i}, \mathrm{Pose}'_{Y_i}, \mathrm{Pose}'_{\theta_i})\}_{i=1}^k$ (Fig \ref{fig:construct}b, top). The spatial attention module, derived from the Convolutional Block Attention Module (CBAM) \cite{woo2018cbam}, is placed at the beginning of the network to emphasize or suppress large-scale geometric features of the input images to help place contact poses within the film boundaries. Our self-supervised SDCNN model is trained on 8,500 augmented images of experimentally synthesized perovskite drop-casted films with an 80/20 training-validation split. The spatially differentiable loss function directly transforms the input predicted poses to the optimization objective in Eq. \ref{eq:loss} to minimize. For the remaining 7 CNNs, a set of labels is generated for the 8,500 training images using a stochastic process. Each label is generated by inputting $N=100$ randomly generated poses into Eq. \ref{eq:loss} for every image. The pose with the lowest loss, $\mathrm{Pose}_{i^*}$, becomes the image label for training:
\begin{equation}
    i^* = \arg \min_{i \in \{1,...,N\}} \; \mathrm{loss}\left( I, \mathrm{Pose}_i \right) \; .
\end{equation}
Although this trial-and-error process is slow, it generates effective data labels for benchmark purposes using the same loss function construction as our SDCNN but without spatial differentiability. Thus, enabling meaningful comparisons between model results. Figure \ref{fig:cnn}c (left) shows an example of the performance of this stochastic process for generating labeled image data.

\subsection*{Path planning experiments}

Path plans are generated by five methods and tested across 115 unique graphs. Each graph consists of new contact poses predicted by the SDCNN using computer vision-segmented images of arrays of 35 drop-casted mixed-halide MAPb(Br$_{1-x}$I$_x$)$_3$ perovskite semiconductor films from different experimental cycles. The five tested algorithms include a genetic algorithm (GA), A*, Christofides algorithm, Dijkstra's algorithm, and our proposed noisy Dijkstra's algorithm. The GA is implemented using the built-in functions available from the \textit{scikit-opt} optimization package in Python \cite{guofei9987_scikitopt}, A* is implemented based on \cite{Hart1968}, Christofides algorithm is implemented based on \cite{christophides1976worst}, and Dijkstra's algorithm is implemented based on \cite{dijkstra1959note}. However, each approach has been slightly modified for solving an OTSP rather than a TSP by relaxing the constraint of the generated route returning to the start node. The GA hyperparameters have been hand-tuned to $1,000$ generations, a population size of $10$, and a mutation rate of $80$\% to produce optimal results on the planning task of minimizing total travel distance. Similarly, the hyperparameters of the proposed noisy Dijkstra's algorithm have been hand-tuned to $1,000$ generations with a noise level of $\alpha=0.02$.

\subsection*{Photoconductivity characterization}

To characterize the photoconductive properties of each perovskite film, a Python-controlled Keithley 2425 source meter measures the resultant current from our SDCNN-driven robotic system with an illuminating four-point probe end effector. The current is measured at 40 unique voltage steps across a -40 V to 40 V voltage sweep for each contact pose to capture detailed current-response curves. This sweep is repeated twice for each contact, once in the dark and once under illumination to measure the photocurrent, $I_\mathrm{ph}$ (Eq. \ref{eq:photo}). Attached to the gold-tipped four-point probe end effector is a 3D-printed LED mount (Fig. \ref{sfig:mount}), which is controlled using Python commands sent to an Arduino microcontroller with solid-state relays. Illumination is provided by probe-mounted high-power white LEDs positioned 4 mm above the film surface. This setup ensures consistent and uniform lighting at an intensity of approximately 200 mW cm$^{-2}$. The 3D-printed LED mount for the probe orients the LEDs to maximize the distribution of uniform light across the measurement area while avoiding shading effects. Additionally, the LEDs have a large viewing angle of 120$^\circ$ to optimize the overlap of light beams, further improving the light distribution uniformity during each measurement.

\section*{Data Availability}

All result files and model weights have been deposited in the OSF database under accession code: \href{https://osf.io/sdy7k}{https://osf.io/sdy7k}.

\section*{Code Availability}

All code used to develop the SDCNN models is available publicly with complete working examples on GitHub: \href{https://github.com/PV-Lab/SDCNN}{https://github.com/PV-Lab/SDCNN}.

{\small
\bibliography{references}
}

\section*{Acknowledgements}

The authors thank Julia Hsu for her contribution to the methodology of this research and Tianran Liu for providing perovskite materials to help calibrate measurements. The authors acknowledge funding support from: First Solar; Eni S.p.A. through the MIT Energy Initiative; University of Toronto’s Acceleration Consortium; and U.S. Department of Energy’s Office of Energy Efficiency and Renewable Energy (EERE) under the Solar Energy Technology Office (SETO) Award Number DE-EE0010503. This work made use of the MRSEC Shared Experimental Facilities at MIT, supported by the National Science Foundation under award number DMR-1419807.

\section*{Author Contributions}

A.E.S. conceptualized the work. A.E.S., B.D., K.J., and T.B. designed the methodology. A.E.S. wrote the software. A.E.S., B.D., and F.S. prepared the experimental materials. A.E.S. and B.D. conducted experiments. A.E.S. performed the analysis. A.E.S. wrote the manuscript. All authors reviewed and edited the manuscript. B.D., K.J., and T.B. provided guidance.

\section*{Competing Interests}

The authors declare no competing interests.

\newpage
\beginsupplement
\onecolumn
\begin{centering}
\large{\textbf{Supplementary Information:}\\A Self-Supervised Robotic System for Autonomous Contact-Based Spatial Mapping of Semiconductor Properties} \par
\end{centering}

\begin{description}
\begin{centering}

\item \textbf{Alexander E. Siemenn}$^{1 * }$, Basita Das$^{1}$, Kangyu Ji$^{1, 2}$, Fang Sheng$^{1}$, Tonio Buonassisi$^{1}$
\item $^{1}$Department of Mechanical Engineering, Massachusetts Institute of Technology, Cambridge, MA 02139, USA
\item $^{2}$Research Laboratory of Electronics, Massachusetts Institute of Technology, Cambridge, MA 02139, USA
\item 
\item $^*$Corresponding author: asiemenn@mit.edu

\end{centering}
\end{description}

\section*{Design}

\noindent \textbf{Robot control workflow}

\begin{figure}[h!]
\begin{center}
\includegraphics[width=1\columnwidth]{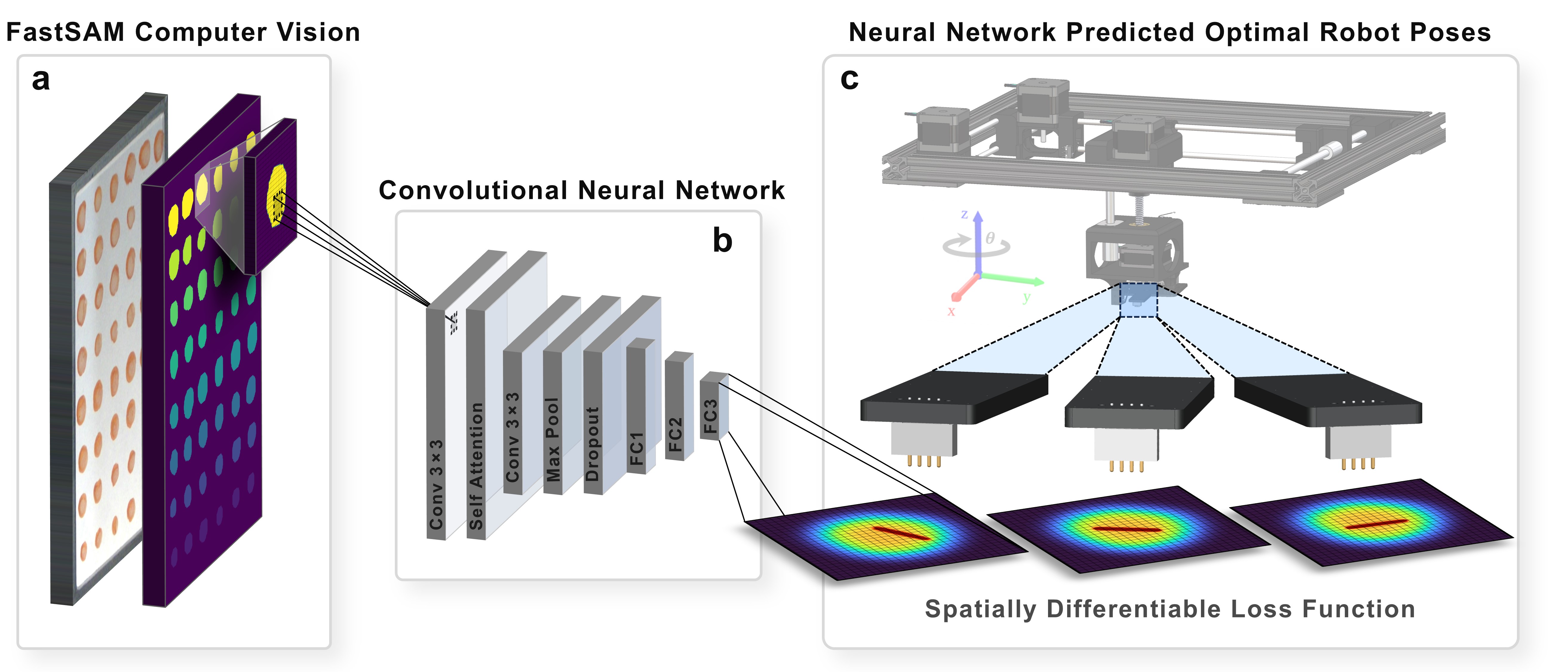}
\end{center}
   \caption{Autonomous robotic workflow, zoomed in. \textbf{a}, The fast segment anything model (FastSAM) is used to quickly segment printed materials. These shapes are used as priors for the spatially differentiable convolutional neural network (SDCNN) predictions. \textbf{b}, SDCNN takes input shape priors, and outputs predicted poses. \textbf{c}, Control of robot motion based on the predicted poses.}
\label{sfig:workflow}
\end{figure}

In this paper, we autonomously control a 4-degree-of-freedom (4DOF) robot with an illuminating four-point probe end effector to measure the photoconductivity of semiconductors. In Fig. \ref{sfig:workflow}, shape priors are used to output optimal poses for the robot to move to and measure each material. A geometric transfer function is used to control the motion of the end effector about the $\theta$-axis (yaw).

\newpage

\noindent \textbf{Robot calibration}

\begin{figure}[h!]
\begin{center}
\includegraphics[width=0.8\columnwidth]{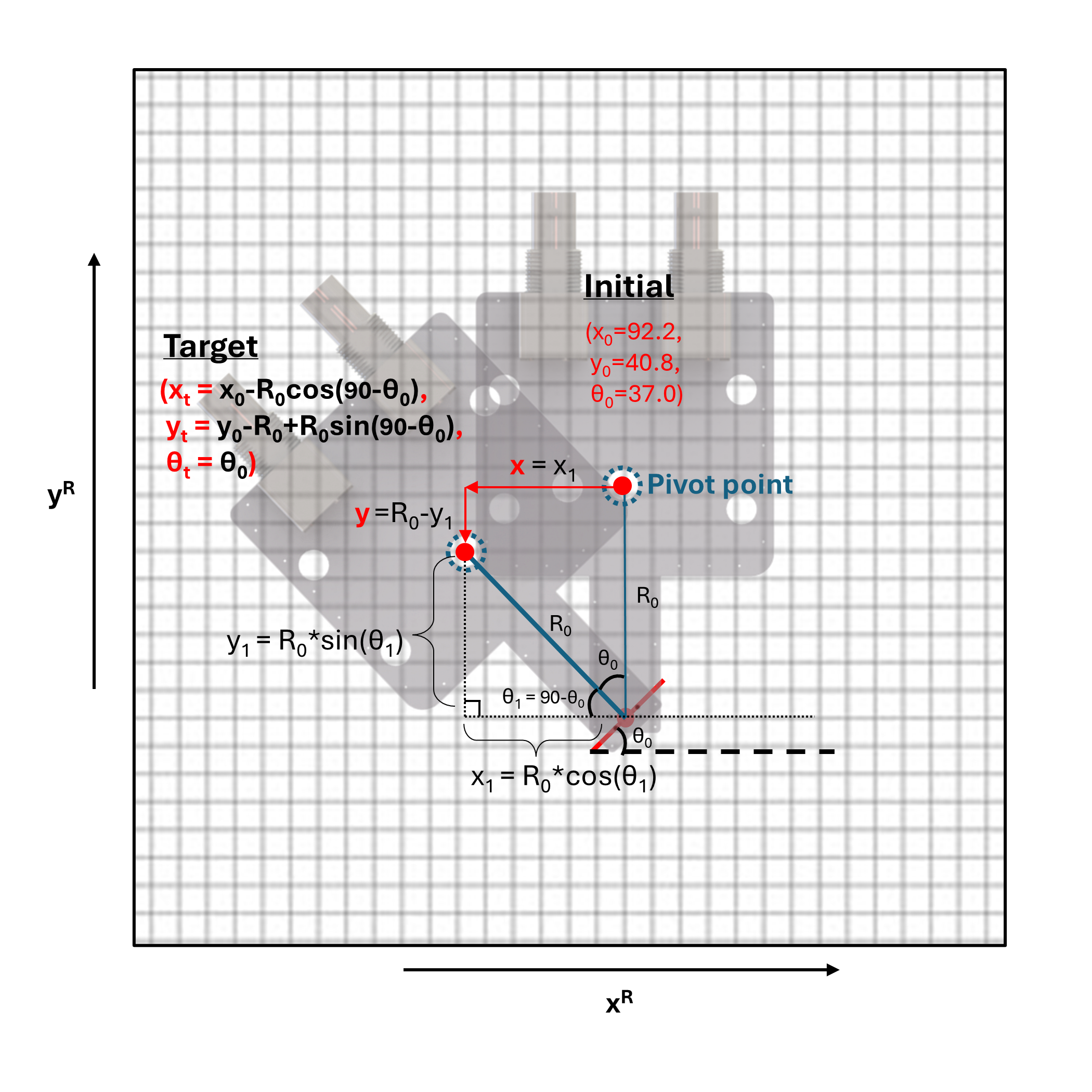}
\end{center}
   \caption{Geometric transfer functions to achieve target poses of an end effector with a single pivot point.}
\label{sfig:geometric}
\end{figure}

Figure \ref{sfig:geometric} illustrates the mathematical procedure of achieving a target pose from an input set of coordinates. A geometric transformation is applied to ensure the contact point down the arm of the end effector meets the coordinates at the correct position in space:
\begin{align}
    x_t &= x_0-R_0\mathrm{cos}(90-\theta_0),\\
    y_t &= y_0-R_0 +R_0\mathrm{sin}(90-\theta_0), \\
    \theta_t &= \theta_0,
\end{align}
where $x_t$, $y_t$, and $\theta_t$ are the target coordinates, given the initial coordinates $x_0$, $y_0$, and $\theta_0$ and $R_0$ is the measured length of the end effector arm from the pivot point to the desired contact point.

\begin{figure}[h!]
\centering
\begin{subfigure}[b]{0.49\textwidth} 
\includegraphics[width=\textwidth]{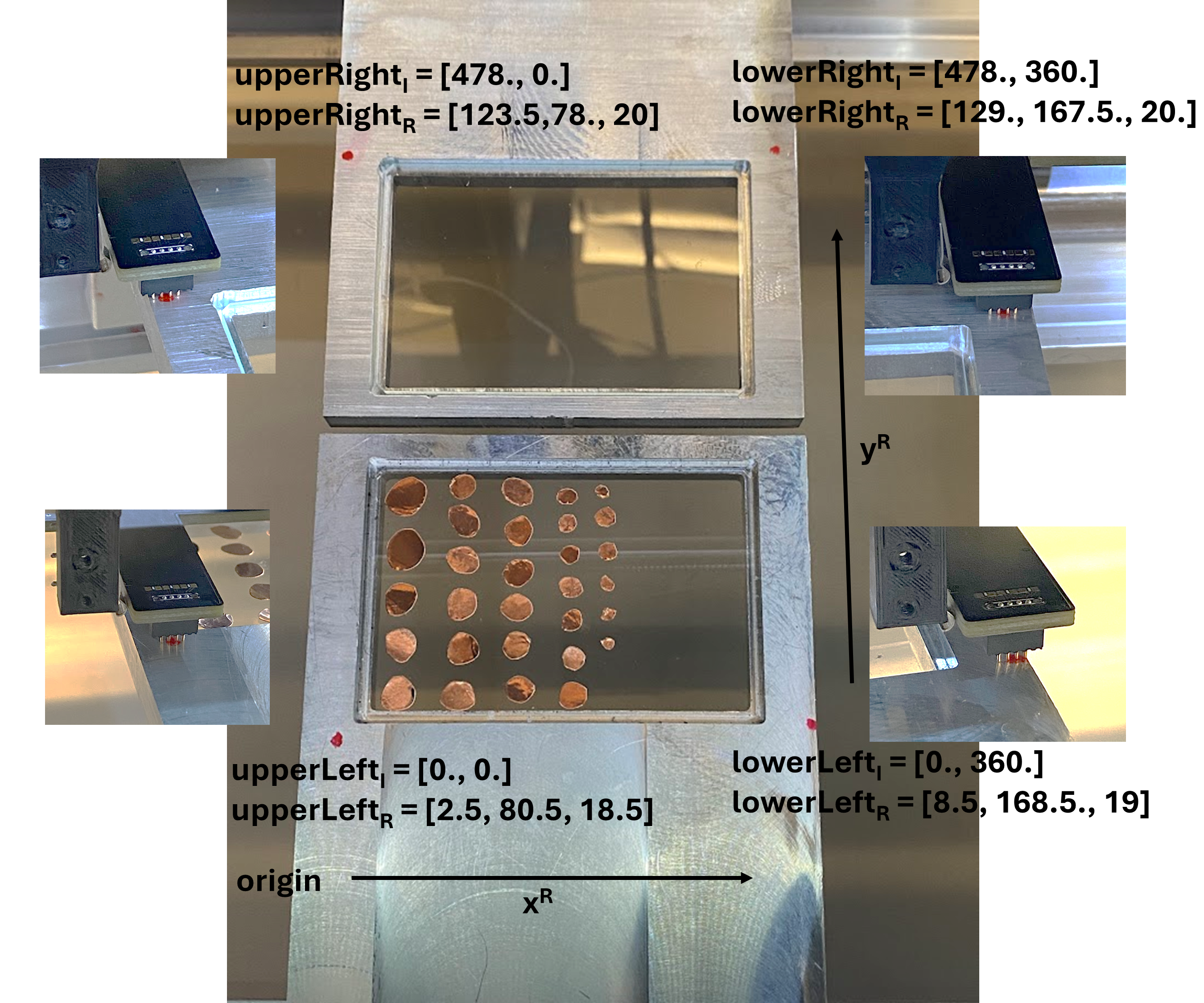}
\caption{Robot coordinate-pair calibration}
\end{subfigure}
\begin{subfigure}[b]{0.49\textwidth} 
\includegraphics[width=\textwidth]{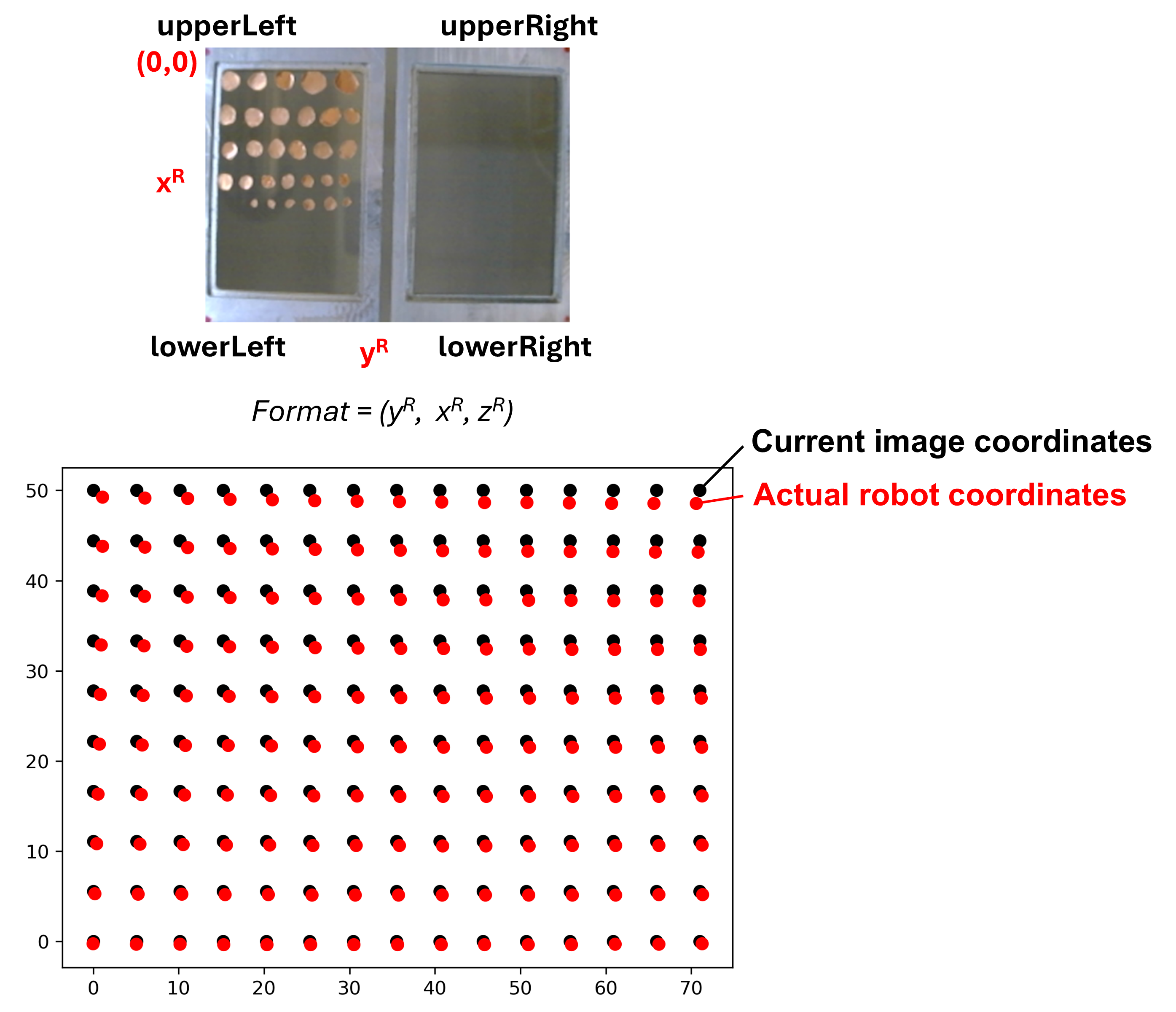}
\caption{Calibration Mesh}
\end{subfigure}
\caption{Calibration procedure for the 4DOF robot. \textbf{a}, Manual calibration of image (I)-robot (R) coordinate pairs. \textbf{b}, Calibration mesh applied to the post-rectified coordinates.}
\label{sfig:calibration}
\end{figure}

To fully calibrate the robot positioning system, image-robot coordinate pairs are collected. These coordinate-pairs establish the robot's location in space, given a set of image coordinates. Coordinate pairs are collected along the outer perimeter of the imaging field. Thus, the corners of the imaging field will align with the corners of the robot reference frame (Fig. \ref{sfig:calibration}a). Due to lens effects and imperfect image rectification, there is still an image-robot coordinate mismatch if measured again after calibration. We measure 15 different image-robot coordinate pairs and generate a calibration mesh (Fig \ref{sfig:calibration}b) to apply to the coordinates to ensure proper alignment of the robot within image space. 

\bigbreak

\noindent \textbf{3D-printed LED mount}

\begin{figure}[h!]
\centering
\begin{subfigure}[b]{0.49\textwidth} 
\includegraphics[width=\textwidth]{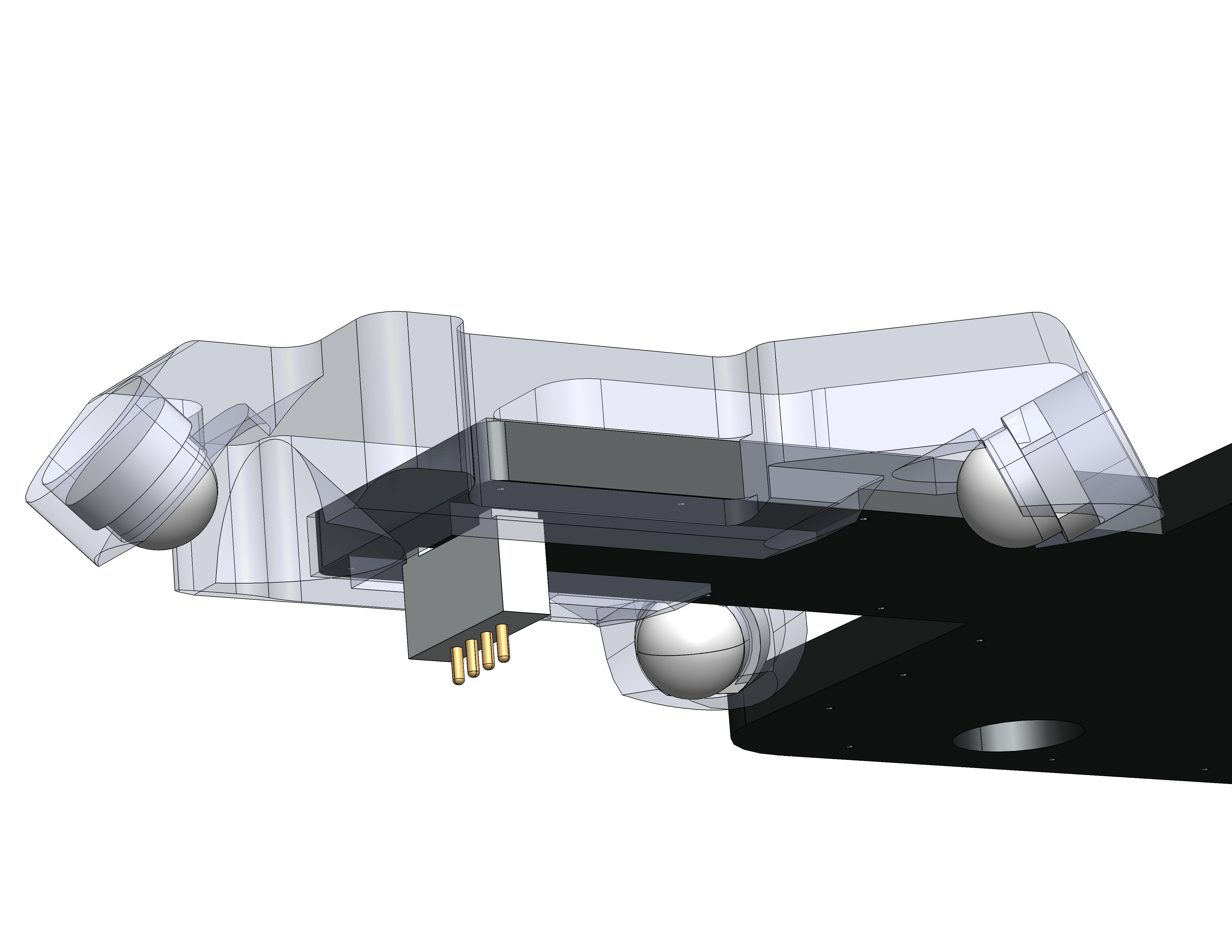}
\caption{Side view of mount}
\end{subfigure}
\begin{subfigure}[b]{0.49\textwidth} 
\includegraphics[width=\textwidth]{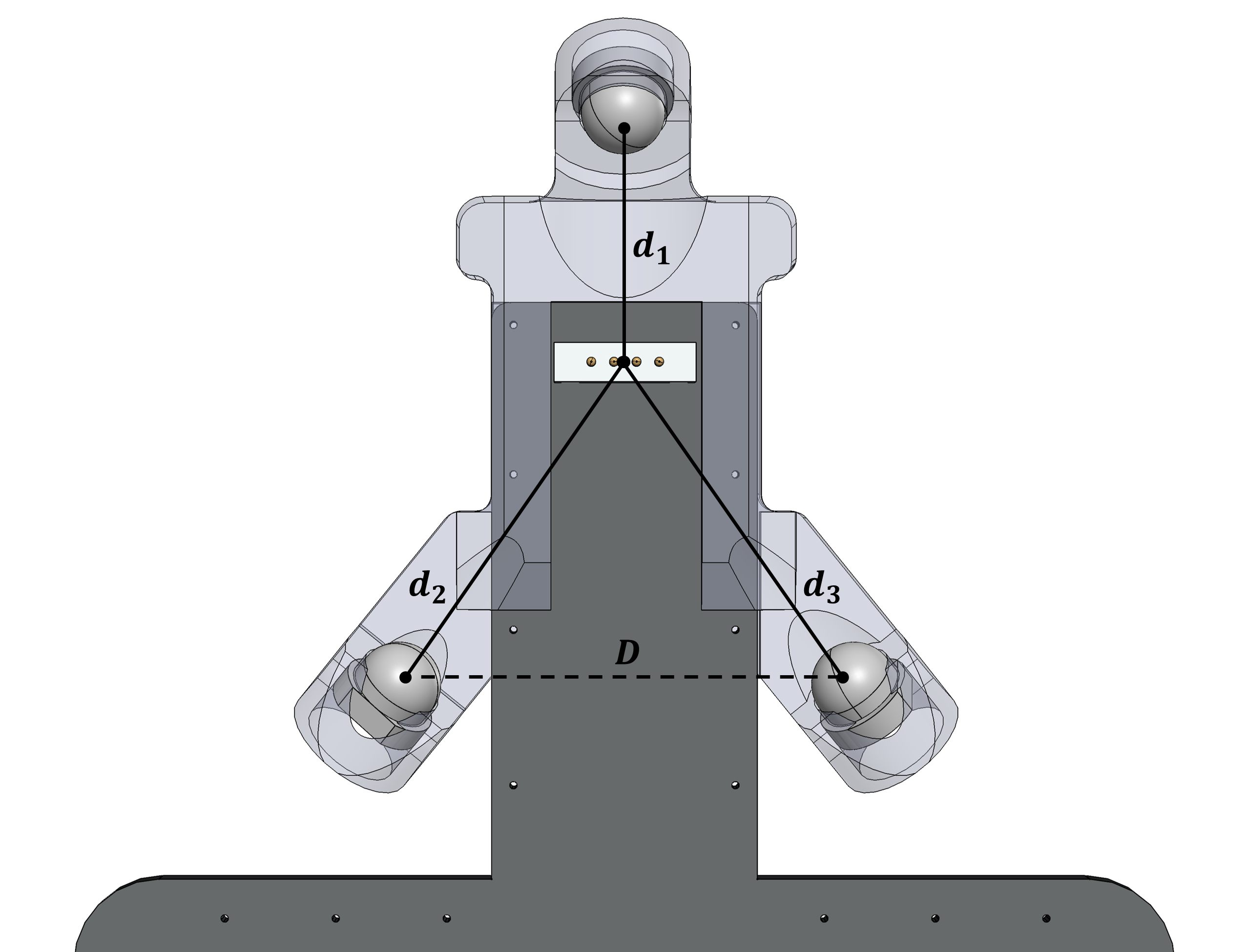}
\caption{Bottom view of mount}
\end{subfigure}
\caption{Design of the 3D-printed LED mount to the four-point probe for photoconductivity measurement. \textbf{a}, Side view of the 3D printed mount. \textbf{b}, Bottom view of the 3D printed mount with calculated LED positions.}
\label{sfig:mount}
\end{figure}

Fuse deposition modeling (FDM) 3D printing is used to create the LED mount for the four-point probe (Fig. \ref{sfig:mount}). Three LEDs are positioned as vertices to an isosceles triangle with the aim of maximizing illumination uniformity at the probe tips and minimizing shading. To minimize shading due to the probe tips, placing two LEDs, one in front and one behind the tips, would suffice. However, it is mechanically infeasible to place an LED behind the tips and still make contact with a material. Thus, we achieve this goal by positioning two LEDs to either side of the probe arm. To ensure uniform light distribution, these back LEDs are mounted farther away from the center point such that illumination intensity is not stronger, using the following relation:
\begin{equation}
    d_3=d_2=\sqrt{(d_1)^2-D^2} \; ,
\end{equation}
where $d_3$ and $d_2$ are the distances from the center point to either back LED, $D$ is the distance between the back LEDs, and $d_1$ is the distance from the center point to the front LED.

\section*{Experiments}

\noindent \textbf{Varying image inputs}

\begin{figure}[h!]
\centering
\begin{subfigure}[b]{0.49\textwidth} 
\includegraphics[width=\textwidth]{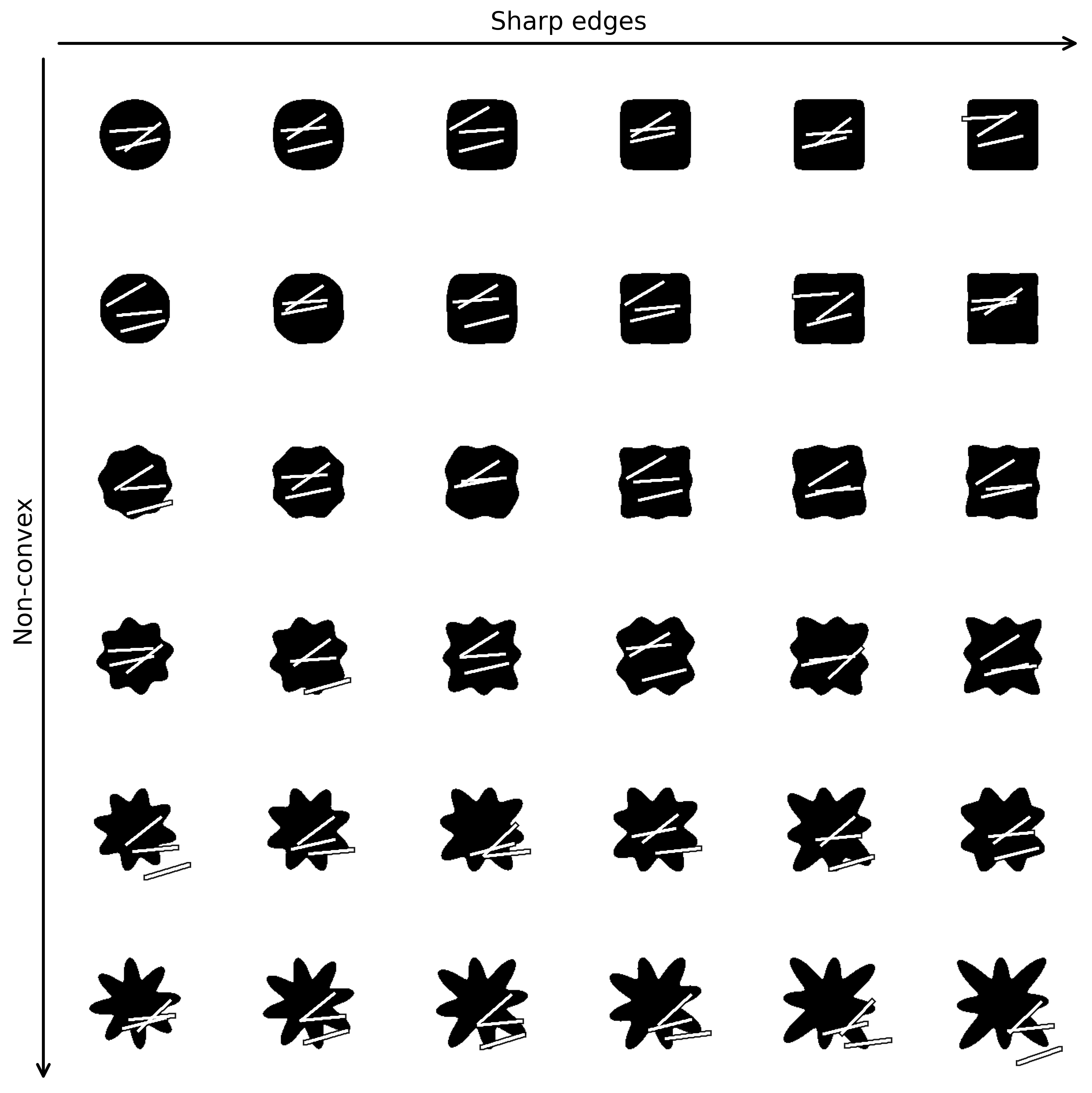}
\caption{Input segments with predictions}
\end{subfigure}
\begin{subfigure}[b]{0.49\textwidth} 
\includegraphics[width=\textwidth]{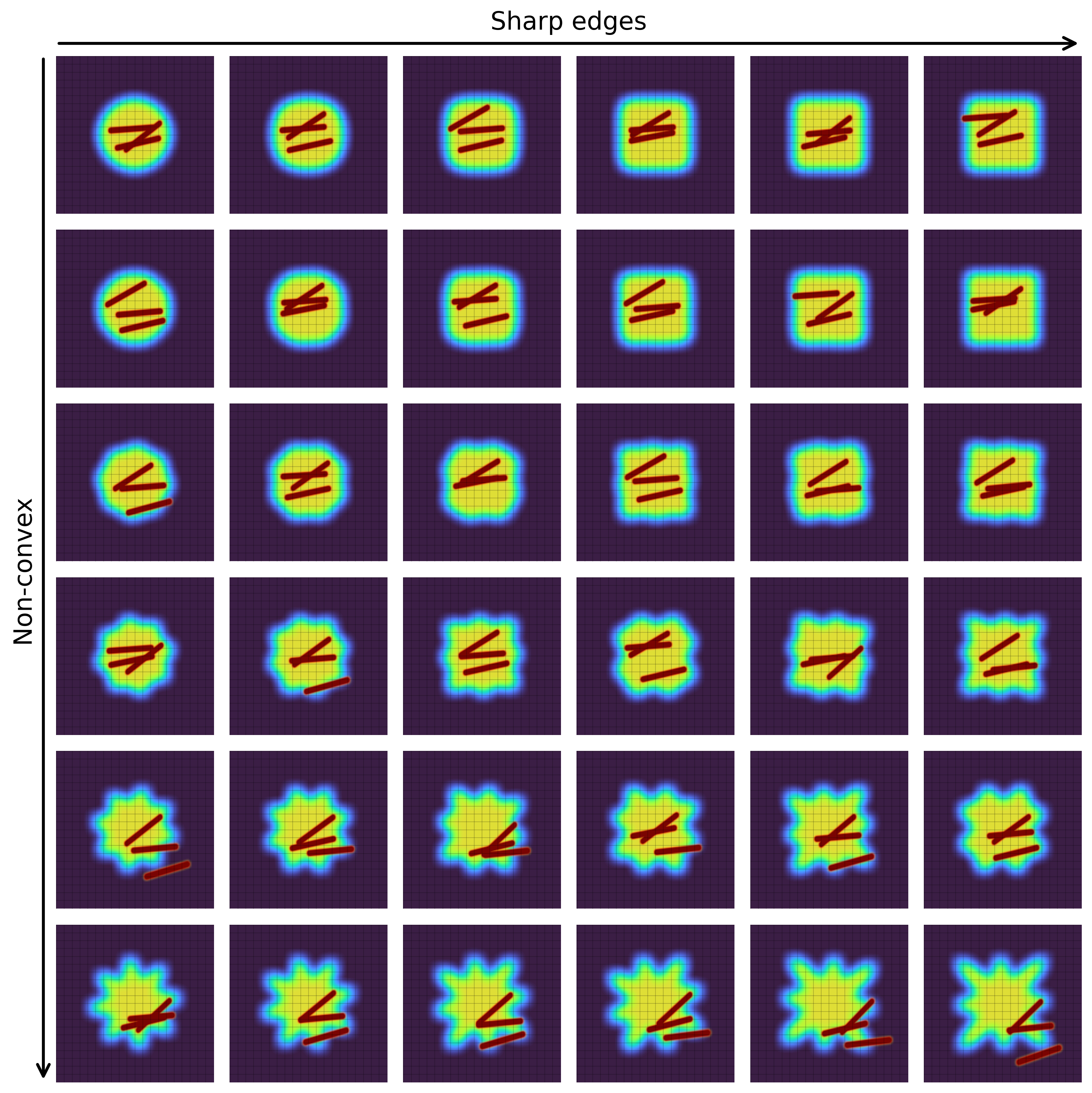}
\caption{Differentiable maps of input shapes}
\end{subfigure}
\caption{SDCNN pose prediction on input shapes varying in edge sharpness and convexity. \textbf{a}, 6$\times$6 grid of various input shapes with predicted poses overlayed. \textbf{b}, 6$\times$6 grid showing the estimated differentiable maps for each input shape.}
\label{sfig:shapes}
\end{figure}

The shape of the input image segments to the SDCNN governs the placement of poses. With the presented version of the model, the weights are trained on rounded, convex shapes due to the nature of predicting poses for drop-casted semiconductor films. However, when we take this pre-trained model and apply it to input shapes with different formats, we get varying results. Figure \ref{sfig:shapes} illustrates how well this SDCNN performs for input shapes varying in edge sharpness and convexity. Model predictions are augmented with $k=3$ poses randomly selected from four predictions for each input to increase randomness and demonstrate robustness to the task. We see that, in general, this SDCNN performs best on convex and rounded input shapes (upper left), as these are most similar to its training set. Robustness to edge sharpness is demonstrated along the horizontal axis, but model performance breaks down as input shapes become more non-convex along the vertical axis. Here, predicted poses start to drift towards the edges, with many falling outside of the shape boundary. These results highlight the robustness of the model to certain features but also show the importance of training the SDCNN using shape priors that more closely align with the expected testing conditions.




\bigbreak
\noindent \textbf{Material characterization}

\begin{figure}[h!]
\begin{center}
\includegraphics[width=0.6\columnwidth]{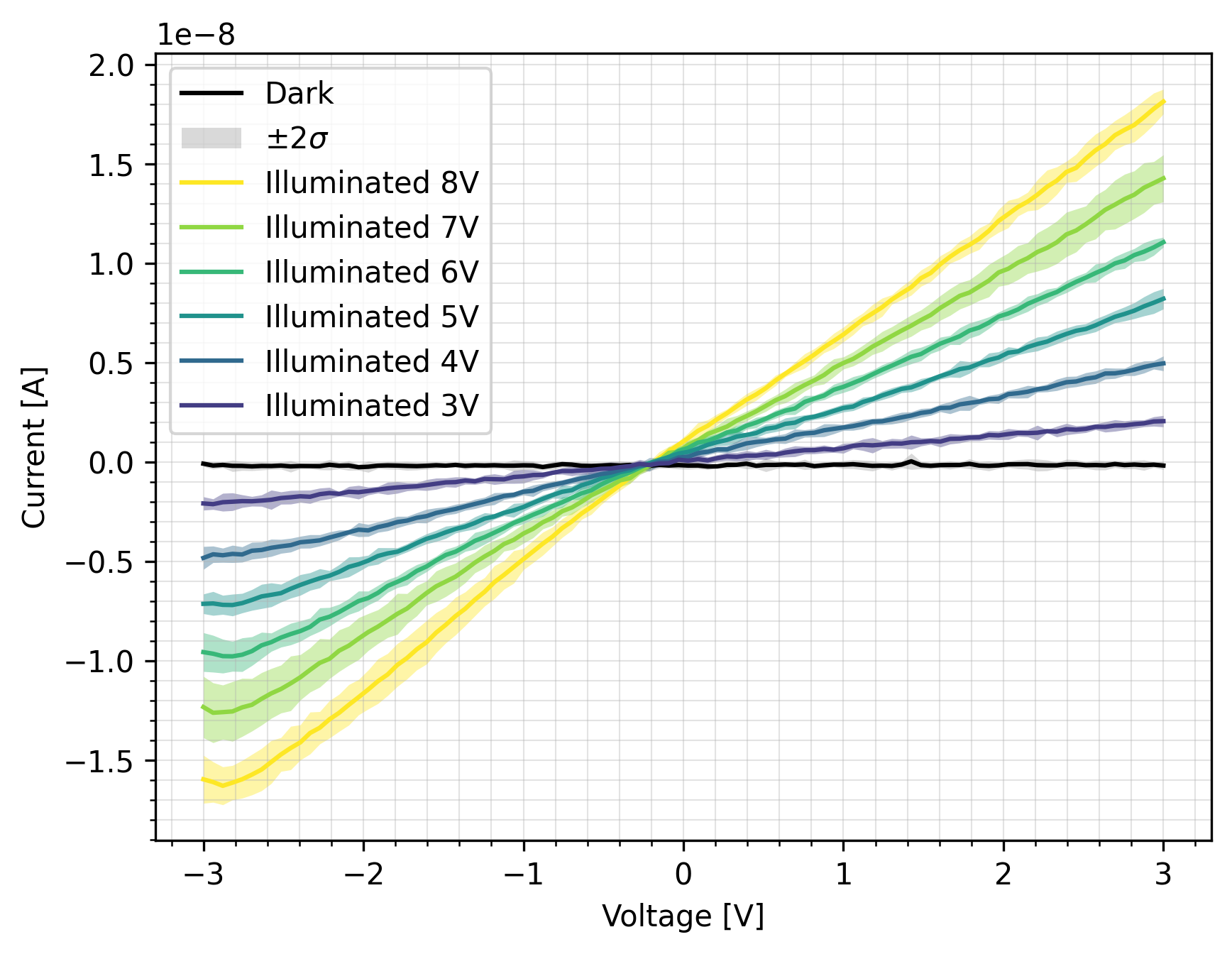}
\end{center}
   \caption{Varying illumination intensity on perovskite photocurrent. Current-voltage response increases as illumination intensity increases for a spun-coat FAPbI$_3$ (formamidinium lead iodide) perovskite thin film. Shaded regions indicate two standard deviations about the mean for each curve across 10 independent measurement trials.}
\label{sfig:fapbi}
\end{figure}

We design the photoconductivity end effector of the 4DOF robot to have variable illumination control. Although we only need dark and illuminated conditions to characterize photoconductance, we calibrate the measurement on all quantized illuminations. Figure \ref{sfig:fapbi} illustrates the current response dependence on illumination intensity for a spun-coat FAPbI$_3$ (formamidinium lead iodide) perovskite thin film. Six different LED illumination intensities are tested, as well as dark conditions. As the voltage supplied to the LEDs increases, the illumination intensity increases. As the illumination intensity increases, the photocurrent response of the FAPbI$_3$ film increases.

\begin{figure}[h!]
\begin{center}
\includegraphics[width=1\columnwidth]{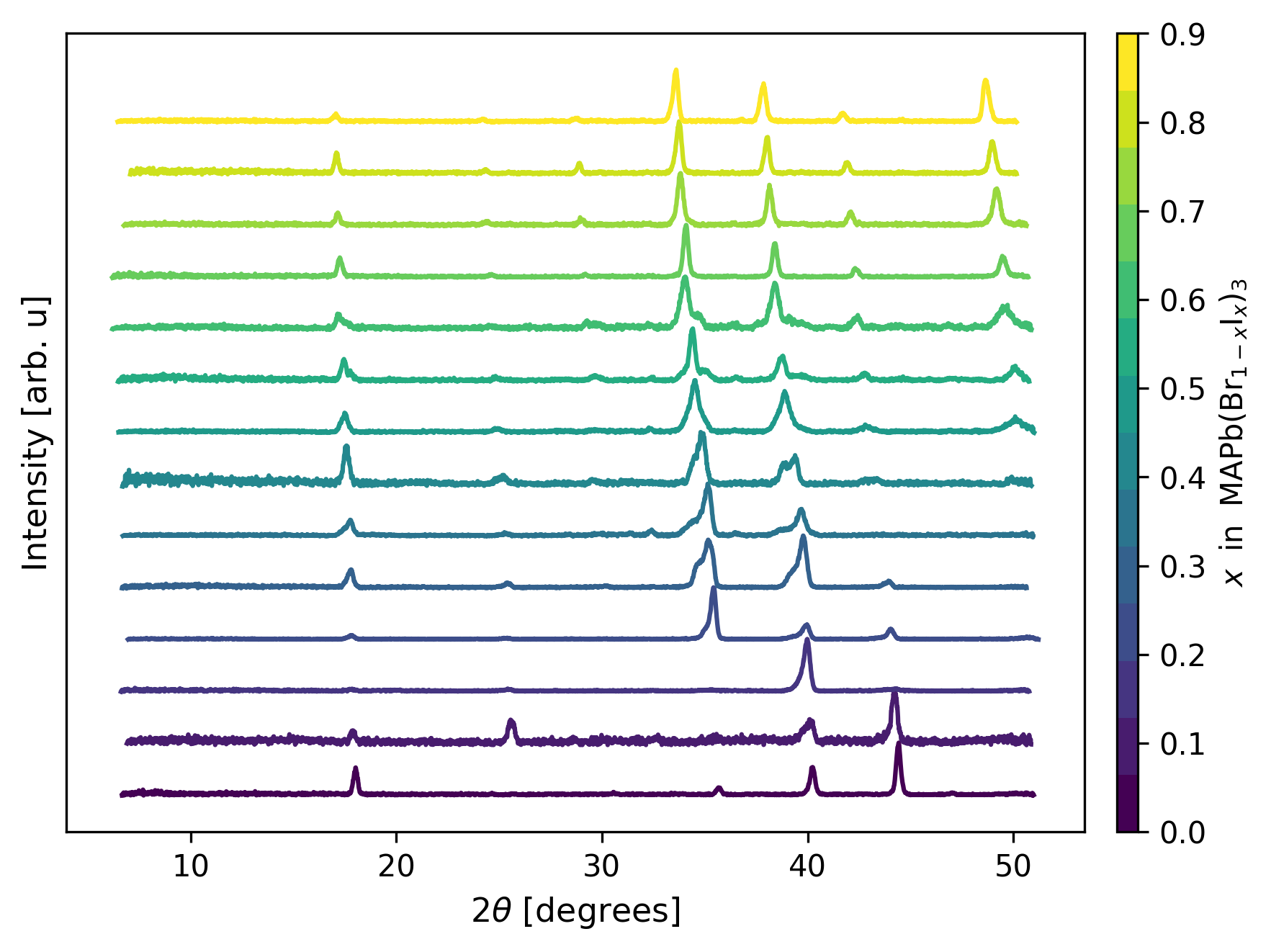}
\end{center}
   \caption{X-ray diffraction (XRD) traces for a gradient of drop-casted MAPb(Br$_{1-x}$I$_x$)$_3$ films using OpenTrons. XRD traces are measured using a Bruker X-ray Diffractometer with a Cobalt Source D8 and General Area Detector Diffraction System.}
\label{sfig:xrd}
\end{figure}

We characterize the crystal phase of the semiconductor films drop-casted by the OpenTrons overhead volumetric pipetter used in this study. Figure \ref{sfig:xrd} illustrates 14 X-ray diffraction (XRD) traces measured from 14 of the total 35 drop-casted methylammonium lead bromide (MAPbBr$_3$) to methylammonium lead iodide (MAPbI$_3$) mixed-halide perovskite semiconductor films used in this study for characterizing photoconductance. We see a clear trend of the XRD peaks shifting along the MAPb(Br$_{1-x}$I$_x$)$_3$ gradient, validating that the perovskites do form a gradient.

\end{document}